%% file: iclr2025_conference.tex
\documentclass{article} 
\usepackage{iclr2025_conference,times}

\input{math_commands.tex}

\usepackage{hyperref}
\usepackage{url}
\usepackage{algorithm}
\usepackage{algorithmic}
\usepackage{graphicx}
\usepackage{colortbl}  
\usepackage{xcolor}    
\usepackage{booktabs}
\usepackage{multicol}
\usepackage{multirow}
\usepackage{newfloat}
\usepackage{listings}
\usepackage{wrapfig}
\usepackage{caption}
\usepackage{enumitem}
\usepackage{booktabs}
\usepackage{tocloft}
\usepackage{marvosym}

\newcommand{\listcasestudyfiguresname}{\normalsize{List of Case Study Figures}}
\newlistof{casestudyfigures}{csf}{\listcasestudyfiguresname}

\newcommand{\casestudyfigure}[4]{%
  \clearpage
  \begin{figure}[ht]
    \centering
    \refstepcounter{casestudyfigures}%
    \addcontentsline{csf}{casestudyfigures}{\protect\numberline{\thecasestudyfigures}#2}%
    \includegraphics[width=0.98\textwidth]{#1}
    \caption{#3}
    \label{#4}
    \hyperlink{listofcasestudyfigures}{Back to List of figures}
\end{figure}}

\title{Can MLLMs Understand the Deep Implication Behind Chinese Images?}

\author{%
    Chenhao Zhang\textsuperscript{1,2}\thanks{Equal Contribution. \scalebox{0.8}{\Letter} \texttt{ch\_zhang@hust.edu.cn};
    \texttt{fengxi@ustc.edu}}\quad
    Xi Feng\textsuperscript{2,3}\footnotemark[1]\quad
    Yuelin Bai\textsuperscript{2}\footnotemark[1]\quad 
    Xinrun Du\textsuperscript{4,5}\footnotemark[1]\\
    \textbf{Jinchang Hou}\textsuperscript{2,3}\enspace
    \textbf{Kaixin Deng}\textsuperscript{6}\enspace
    \textbf{Guangzeng Han}\textsuperscript{7}\enspace
    \textbf{Qinrui Li}\textsuperscript{8}\enspace 
    \textbf{Bingli Wang}\textsuperscript{9}\enspace
    \textbf{Jiaheng Liu}\textsuperscript{4}\\
    \textbf{Xingwei Qu}\textsuperscript{10}\enspace
    \textbf{Yifei Zhang}\textsuperscript{11}\enspace 
    \textbf{Qixuan Zhao}\textsuperscript{2,3}\enspace 
    \textbf{Yiming Liang}\textsuperscript{12}\enspace 
    \textbf{Ziqiang Liu}\textsuperscript{2}\enspace
    \textbf{Feiteng Fang}\textsuperscript{2,3}\\
    \textbf{Min Yang}\textsuperscript{2}\enspace
    \textbf{Wenhao Huang}\textsuperscript{5}\enspace
    \textbf{Chenghua Lin}\textsuperscript{11}\enspace 
    \textbf{Ge Zhang}\textsuperscript{4,5}\thanks{Corresponding authors. \scalebox{0.8}{\Letter} \texttt{gezhang@umich.edu};
    \texttt{sw.ni@siat.ac.cn}}\quad
    \textbf{Shiwen Ni}\textsuperscript{2}\footnotemark[2]\\
    \textsuperscript{1}Huazhong University of Science and Technology\enspace
    \textsuperscript{2}Shenzhen Institute of Advanced Technology, CAS\\
    \textsuperscript{3}University of Science and Technology of China\enspace
    \textsuperscript{4}M-A-P\enspace
    \textsuperscript{5}01.ai\enspace
    \textsuperscript{6}CDUT\enspace
    \textsuperscript{7}University of Memphis\\
    \textsuperscript{8}University of California, Santa Barbara\enspace
    \textsuperscript{9}SICAU\enspace
    \textsuperscript{10}University of Manchester\enspace
    \textsuperscript{12}SWU\enspace
    \textsuperscript{12}UCAS
}

\iclrfinalcopy 
\begin{document}

\maketitle
\vspace{-0.6cm}
\begin{abstract}

As the capabilities of Multimodal Large Language Models (MLLMs) continue to improve, the need for higher-order capability evaluation of MLLMs is increasing. However, there is a lack of work evaluating MLLM for higher-order perception and understanding of Chinese visual content.
To fill the gap, we introduce the \textbf{C}hinese \textbf{I}mage \textbf{I}mplication understanding \textbf{Bench}mark, \textbf{CII-Bench}, which aims to assess the higher-order perception and understanding capabilities of MLLMs for Chinese images. 
CII-Bench stands out in several ways compared to existing benchmarks. Firstly, to ensure the authenticity of the Chinese context, images in CII-Bench are sourced from the Chinese Internet and manually reviewed, with corresponding answers also manually crafted. Additionally, CII-Bench incorporates images that represent Chinese traditional culture, such as famous Chinese traditional paintings, which can deeply reflect the model's understanding of Chinese traditional culture.
Through extensive experiments on CII-Bench across multiple MLLMs, we have made significant findings. 
Initially, a substantial gap is observed between the performance of MLLMs and humans on CII-Bench. The highest accuracy of MLLMs attains 64.4\%, where as human accuracy averages 78.2\%, peaking at an impressive 81.0\%. Subsequently, MLLMs perform worse on Chinese traditional culture images, suggesting limitations in their ability to understand high-level semantics and lack a deep knowledge base of Chinese traditional culture. Finally, it is observed that most models exhibit enhanced accuracy when image emotion hints are incorporated into the prompts.
We believe that CII-Bench will enable MLLMs to gain a better understanding of Chinese semantics and Chinese-specific images, advancing the journey towards expert artificial general intelligence (AGI).
Our project is publicly available at \url{https://cii-bench.github.io/}.

\vspace{-0.1cm}
\begin{figure*}[!hbtp]
  \centering  
  \includegraphics[width=0.485\textwidth]{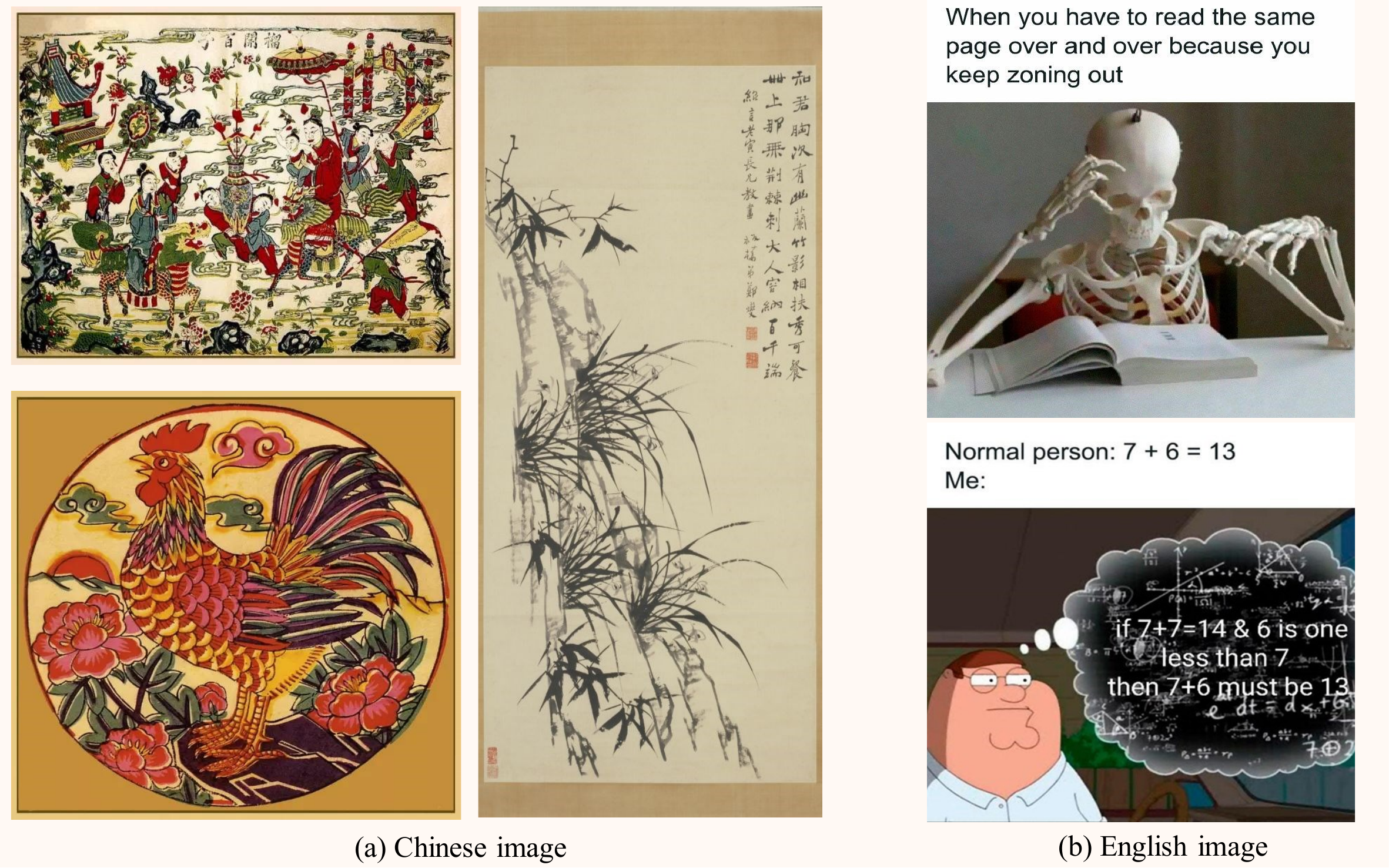}
  \caption{Comparision of Chinese and English image implications. Chinese images often embody richer scenes and deeper implications with Chinese traditional culture compared with the straightforward and explicit symbolism in English images.}
  \label{figure:comparision of image}
\end{figure*}
\end{abstract}

\section{Introduction}

With the rapid advancement of artificial intelligence, Multimodal Large Language Models (MLLMs) \citep{liu2023llava,li2023blip,ye2023mplug, tong2024cambrian1fullyopenvisioncentric} have demonstrated exceptional performance across various domains, including natural language processing \citep{chowdhary2020natural,luo2024graphinstructempoweringlargelanguage,zhang-etal-2024-cpsycoun} and computer vision \citep{lu2022learn,li2023seed,li2023seed2,xu2023lvlm,fu2023mme,cai2023benchlmm,zhang2023m3exam,chen2024right,jin-etal-2024-mmtom}. These models are not only capable of processing and generating text but also excel at integrating and interpreting information across multiple modalities, such as images, sound, and video. However, despite the significant progress made in tasks like image recognition and generation, a crucial research question remains: Can these models truly understand and interpret images that have deep implications? \citep{liu2024iibenchimageimplicationunderstanding} construct an English image implication understanding dataset, II-Bench, and the experiments on MLLMs and human subjects reveal a substantial gap in the models' higher-order perception abilities, particularly in nuanced emotional understanding and profound meaning extraction, when compared to humans.
Unfortunately, the rapid advancement of MLLMs has led to significant performance improvements. For instance, Claude-3.5-Sonnet has achieved an impressive accuracy of 80.9\% on II-Bench, approaching the average human accuracy of 90.3\%. This progress underscores the need for more challenging benchmarks that incorporate richer scenes and deeper implications to continue pushing the boundary of image implication understanding task. 

In contrast to English images, Chinese images often embody richer scenes \citep{10.3389/fpsyg.2023.1198265} and deeper implications as Figure~\ref{figure:comparision of image} shows. For instance, Chinese traditional landscape paintings not only depict natural scenery but also convey profound philosophical concepts, such as the harmony between humans and nature, through artistic techniques like the interplay of void and solid, the use of negative space, and the brushwork. As the famous Chinese poet Su Shi noted, ``Poetry and painting share the same essence, embodying both craftsmanship and purity". The depth of Chinese images lies not only in their aesthetic appeal but also in the underlying spirit and philosophy they express. Similarly, New Year paintings, as a significant carrier of Chinese traditional culture, typically use symbolism and implication to convey wishes for good fortune, prosperity, and peace. Unlike the directness often found in English imagery, Chinese images emphasize the creation of atmosphere and subtle expression, requiring viewers to possess certain cultural knowledge to accurately grasp their meanings. This cultural disparity leads to significant differences in the modes of expression and meaning conveyed between Chinese and English images, highlighting the need to consider cultural context when evaluating the capability of MLLMs to understand the deep implications of images.

To address this gap, we develop CII-Bench, a benchmark designed to comprehensively test the higher-order perception, reasoning, and understanding abilities of models within a Chinese context. This benchmark allows us to gain a clearer understanding of these models' interpretive capacities, offering new insights into their application in cross-cultural environments, and thus advancing the research and development of MLLMs.

\begin{wrapfigure}{r}{0.48\linewidth}
  \centering
  \includegraphics[width=0.48\textwidth]{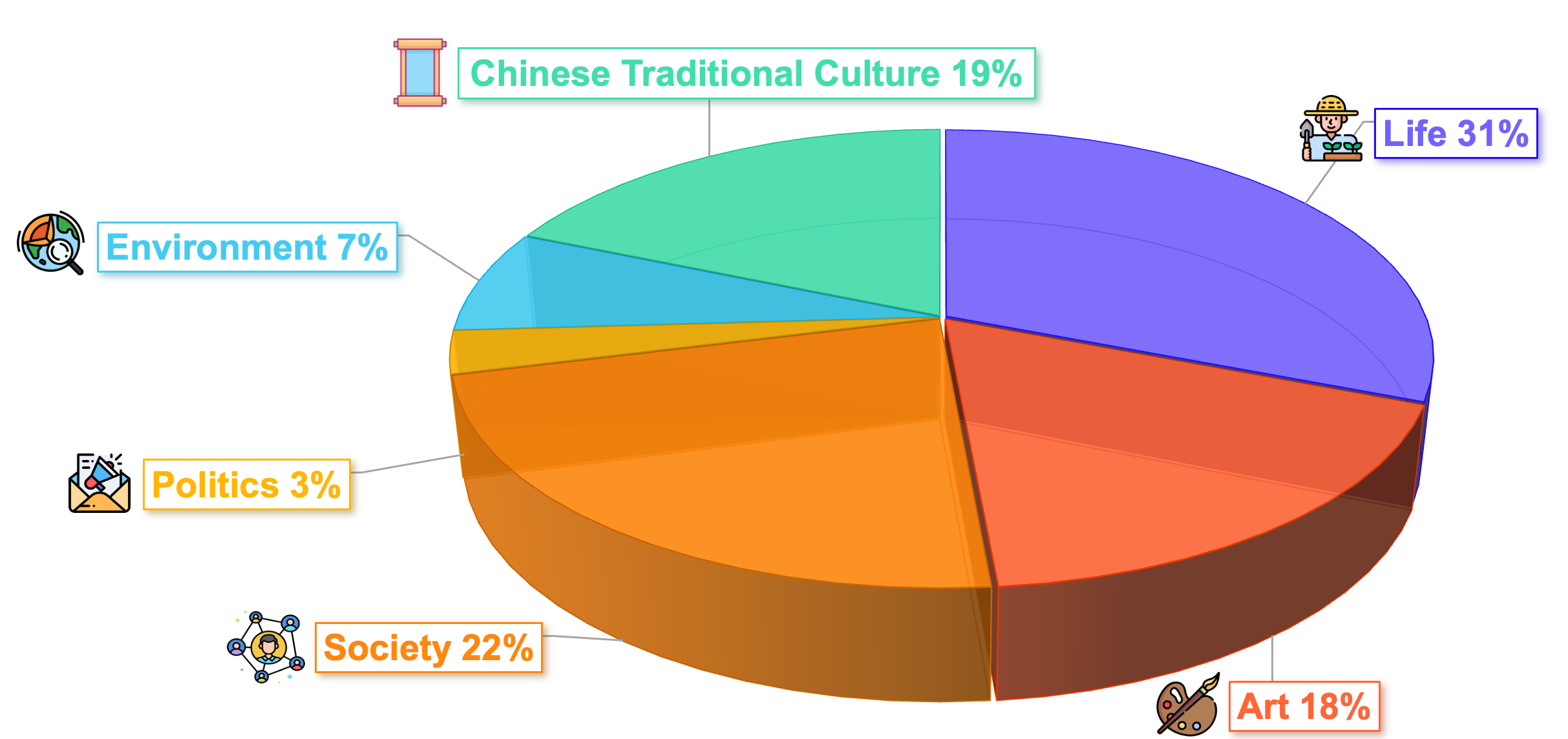}
  \caption{Composition of {CII-Bench}.}
  \label{fig:composition}
\end{wrapfigure}

As illustrated in Figure \ref{fig:composition}, CII-Bench comprises 698 images and 800 multiple-choice questions spanning six domains: Life, Art, Society, Politics, Environment, and Chinese Traditional Culture. Moreover, to ensure diversity, CII-Bench includes six types of images: Illustration, Meme, Poster, Single-panel Comic, Multi-panel Comic, and Painting. By employing images of various types and from different domains, the benchmark provides a more robust evaluation of models' comprehension and reasoning abilities.

We conduct extensive experiments to evaluate CII-Bench on MLLMs that support Chinese and deeply evaluate the model's grasp of Chinese traditional culture. Our key contributions are as follows:

\begin{itemize}[left=10pt]
\item We introduce CII-Bench, the first benchmark designed to assess the understanding of implications in Chinese images, which poses a significant challenge to current MLLMs.
\vspace{-3pt}
\item We design a comprehensive evaluation metric based on GPT-4o to evaluate Chinese traditional culture. This metric aligns more closely with human annotations and is better suited for evaluating Chinese traditional painting. 
\vspace{-3pt}
\item Our experimental findings are as follows:
(1) There is a notable performance gap between MLLMs and humans. Models demonstrate the highest accuracy of 64.4\%, while human accuracy average at 78.2\% and best at 81.0\%.
(2) Closed-source models generally outperform open-source models, but the best-performing open-source model surpasses the top closed-source model, with a difference of more than 3\%.
(3) Models perform significantly worse in Chinese traditional culture compared to other domains, indicating that current models still lack sufficient understanding of Chinese culture. Further analysis shows that GPT-4o can only observe the surface-level information, it's difficult to deeply interpret the complex cultural elements contained in Chinese traditional painting.
(4) Incorporating image emotion hints into prompts generally improves model scores, indicating that models struggle with emotional understanding, leading to misinterpretation of the implicit meanings in the images.

\end{itemize}

\section{Related Work}
\subsection{Multimodal Large Language Models}

With the rapid development of large language models (LLMs) \citep{chowdhery2022palm,chung2022scaling,chiang2023vicuna,touvron2023llama,chatgpt,openai2023gpt,geminiteam2024gemini,cai2024internlm2}, Multimodal Large Language Models (MLLMs) have made significant improvements. Many works incorporate additional module inputs on LLMs, effectively bridging the gap between visual and language. 
BLIP-2~\citep{li2023blip} encodes images using ViT~\citep{vit} and employs a Q-Former to map visual features into the language space. LLaVA \citep{liu2023llava,liu2023improved,liu2024llavanext,li2024llavaonevisioneasyvisualtask} utilizes an MLP as the connector between the visual encoder and the LLM backbone. Similarly, mPLUG-Owl2 \citep{ye2023mplug} employs a modality-adaptive module to facilitate the collaboration between visual and language modalities by mapping them into a unified representation space.
Subsequent works \citep{wang2023cogvlm,lu2024deepseekvl,chen2024far,young2024yi,laurenccon2024matters,glm2024chatglm,yao2024minicpm,Claude3S,Qwen2VL}further enhance MLLMs by designing novel modules for more sufficient modality alignment.

\subsection{MLLM Benchmarks}

The rapid advancement of MLLMs has emphasized the critical need for comprehensive evaluation frameworks within the research community. Initial benchmarks primarily focused on specific tasks, such as visual question answering (VQA) \citep{Antol_2015_ICCV,goyal2017making,kafle2017analysis,singh2019towards,hudson2019gqa} and image captioning \citep{lin2014microsoft,agrawal2019nocaps,plummer2015flickr30k}. While these benchmarks have yielded significant insights, they fall short in providing a holistic assessment of MLLMs across the broader spectrum of multimodal perception and reasoning capabilities.
To address this limitation, recent studies have developed more comprehensive evaluation approaches \citep{xu2023lvlm,fu2023mme,lu2022learn,cai2023benchlmm,zhang2023m3exam,he2024cmmu,chen2024right}. For instance, MMBench \citep{liu2023mmbench} and SEED \citep{li2023seed,li2023seed2} assess models' capabilities through common-sense questions, employing multiple-choice formats to evaluate various dimensions of ability. To assess specialized expertise, MMMU \citep{yue2023mmmu} and CMMMU \citep{zhang2024cmmmu} utilize content derived from exams and textbooks, enhancing the evaluation of domain-specific knowledge. Furthermore, Cambrian-1 \citep{tong2024cambrian1fullyopenvisioncentric} introduces a novel vision-centric benchmark (CV-Bench) to repurpose standard vision tasks for multimodal evaluation.

\subsection{Image Implication Understanding}

Image implication understanding represents a more complex and challenging task than conventional image understanding. This advanced cognitive process necessitates multi-hop reasoning ability and sophisticated theory of mind (ToM), capabilities that are intrinsic to human cognition \citep{Desai_Chakraborty_Akhtar_2022,hessel-etal-2023-androids,yang2024large,zhong2024lets, strachan2024testing,street2024llms,horvitz-etal-2024-getting}. II-Bench \citep{liu2024iibenchimageimplicationunderstanding} is the first benchmark specifically designed to evaluate MLLMs in both image understanding and reasoning through English image implication.

\section{The CII-Bench}
\subsection{Overview of CII-Bench}

We present the \textbf{C}hinese \textbf{I}mage \textbf{I}mplication Understanding \textbf{Bench}mark (CII-Bench), a novel benchmark designed to assess the perceptual, reasoning, and comprehension abilities of MLLMs in the context of Chinese imagery. This benchmark includes a diverse range of visual content such as traditional Chinese traditional artworks, comics, posters, and Chinese Internet memes, all rich in visual information and cultural significance. The main goal of CII-Bench is to evaluate if current MLLMs can leverage their understanding and knowledge of Chinese culture to accurately interpret the deeper implications and abstract information within these images.

\begin{figure*}[!t]
  \centering  
  \includegraphics[width=1.0\textwidth]{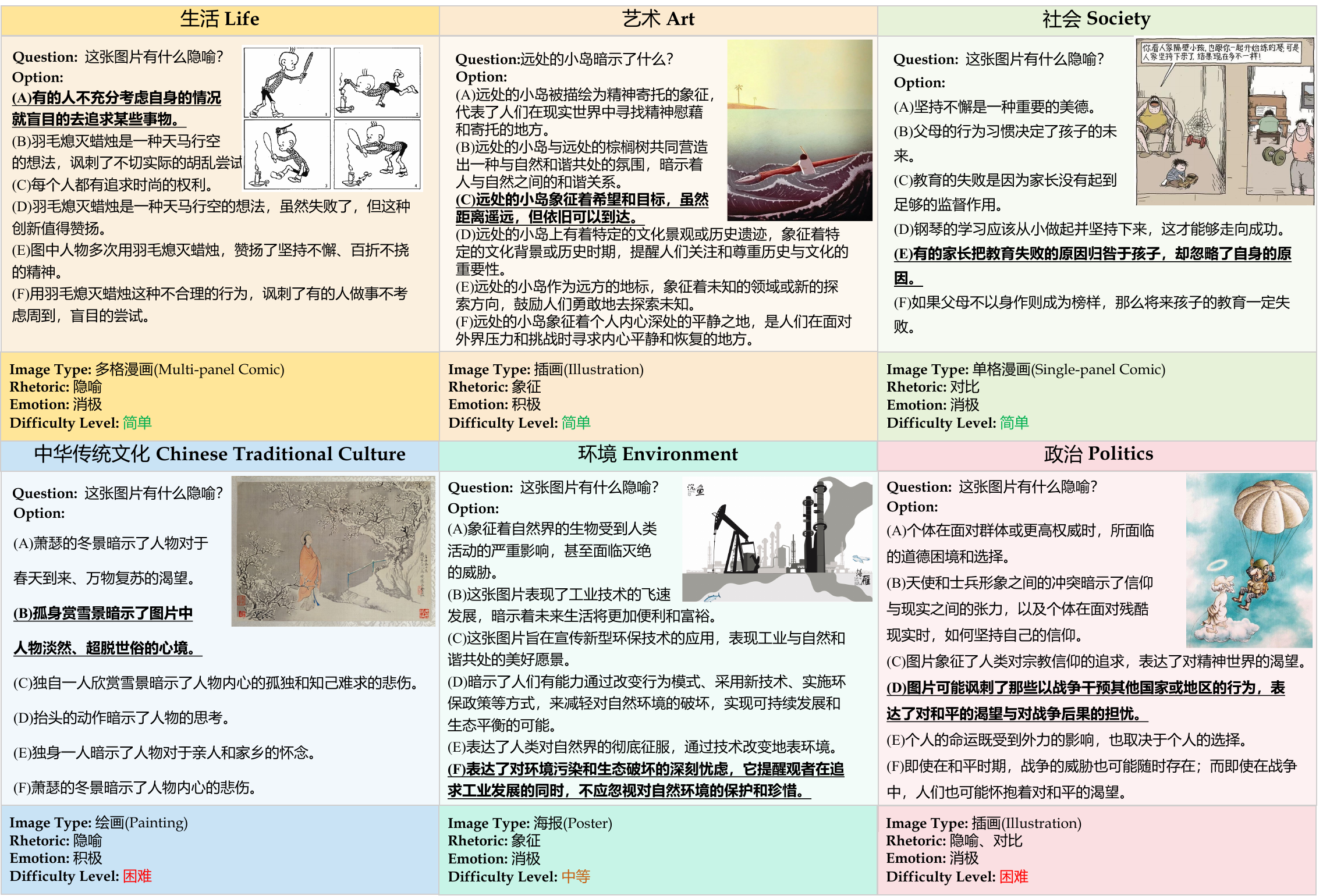}
  \caption{CII-Bench examples sampled from each domain. The English version in Appendix~\ref{sec:appendix-examples}.}
  \label{figure:questions}
\end{figure*}

CII-Bench comprises 698 images across various categories, with detailed classification and domain statistics provided in Appendix \ref{sec:appendix-Data-inf}. These images are manually collected and annotated by 30 undergraduate students from different disciplines and institutions, sourced from several well-known image websites. Each image is paired with 1 to 3 multiple-choice questions, each offering six options with only one correct answer. One fixed question asks, “What is the implication in this image?” Additional questions for the same image probe different levels of understanding, such as overarching interpretation and nuanced details. The benchmark includes 800 multiple-choice questions, with 765 for the test set and 35 for developing and validating few-shot tasks. Figure \ref{figure:questions} provides representative examples from CII-Bench.

\subsection{Data Curation Process}
\subsubsection{Data Collection}

We collect 17,695 raw images from various renowned illustration websites, ensuring a sufficiently extensive raw dataset. Our collectors are well instructed to adhere to copyright and license regulations, avoiding data from sites prohibiting copy and redistribution. For detailed information on the specific websites from which we collect images, please refer to Appendix~\ref{sec:appendix-Data Annotation}.

\subsubsection{Data Filtration}

After collecting the raw images, we meticulously design a three-stage data filtering process: 
In the first stage, we focus on image deduplication. We utilize image similarity algorithms for pixel-level comparison to eliminate duplicates and preserve dataset uniqueness; 
In the second stage, we regulate text prevalence in images. Optical Character Recognition (OCR) technology identifies textual areas and disqualifies images exceeding set text-area ratios, maintaining a visual-centric dataset;
In the third stage, images undergo rigorous visual inspection, discarding those without metaphorical depth based on strict criteria. This process refines the dataset, rejecting over 95\% of initial images and securing under 1,000 high-quality ones.

\subsubsection{Data Annotation}

The annotation process for the benchmark was meticulously designed through several steps to ensure rigor and precision as following. The detailed annotation protocol can be found in Appendix~\ref{sec:appendix-Data Annotation}.

\textbf{Preparation and Consistency Check:} Before formal annotation, annotators first acquaint themselves with standard templates and guidelines. A pre-annotation round on a shared image batch ensures uniform standard understanding, with discrepancies resolved through discussion.

\textbf{Multiple Rounds of Annotation and Cross-Validation:} To reduce bias, each image receives annotations from two different annotators. Cross-validation follows, with a third-party review for significant discrepancies, guaranteeing accuracy.

\textbf{Refinement of Annotation Content:} Annotators annotate each image's difficulty, type, emotional label, domain, and rhetorical devices based on specific criteria, ensuring consistency and comparability.  They also craft 1 to 3 refined questions per image, each with one correct answer among five distractor options, including the default question, ``What is the implication in this image?”

\textbf{Context Analysis:} During the annotation process, annotators assess the image's cultural and background significance, especially for implications and rhetorical devices, consulting relevant materials for accuracy.

\textbf{Post-Annotation Review:} Upon completion, annotations undergo a thorough quality review for any oversight, errors, or inconsistencies. Based on the evaluation results, feedback is provided to the annotators, with re-annotations as necessary to maintain data quality.



\subsection{Dataset Statistics}

CII-Bench comprises 698 images, each accompanied by 1 to 3 multiple-choice questions, totaling 800 questions. We randomly select 35 of these questions to construct a few-shot development set and validation set. On average, each question is approximately 11 characters long, while each option has an average length of 28 characters. Additionally, each image is supplemented with a manually written description by the annotators, which provides a detailed explanation of the image’s content, nuances, and the human interpretation of its deep implication.

CII-Bench covers images across six distinct domains: Life, Art, Society, Politics, Environment, and Chinese Traditional Culture. The types of images are diverse, including Illustration, Meme, Poster, Single-panel Comic, Multi-panel Comic, and Painting. Based on human understanding, these images are categorized into three levels of difficulty: Easy, Medium, and Hard. Moreover, the images are classified according to the emotional information they convey: Positive, Neutral, or Negative. Each image is also manually annotated with the rhetorical devices employed, including Metaphor, Exaggeration, Symbolism, Visual Dislocation, Antithesis, Analogy, Personification, and Contrast. Detailed statistical information is provided in Appendix~\ref{sec:appendix-Data-inf}.

\section{Experiment}

We conduct systematic experiments on both open-source and closed-source MLLMs using CII-Bench. For each model, we employ eight different configurations: None (zero-shot), 1-shot, 2-shot, 3-shot, CoT, Domain, Emotion, and Rhetoric. ``None" represents the use of a standard prompt without any additional information. ``Emotion" indicates the inclusion of information related to the emotional polarity of the image (e.g., positive, negative) in the prompt, ``Domain" involves adding information about the image’s domain (e.g., life, art), and ``Rhetoric" refers to including details about the rhetorical devices used in the image (e.g., metaphor, contrast) in the prompt. Additionally, to verify the necessity of images in problem-solving, we select a portion of LLMs to complete tasks without image input. For consistency across all MLLMs and LLMs, we use identical prompts and experiment setup, with specific details available in Appendix~\ref{sec:appendix-experiments_baseline}. 

\subsection{Baselines}

\textbf{MLLMs.}
To comprehensively evaluate CII-Bench, we carefully select a diverse range of MLLMs, encompassing both open-source and closed-source models, with the aim of covering a wide spectrum of model characteristics and scales. These models span parameter sizes from 7B to 100B, ensuring that models of varying complexity and capability are thoroughly assessed. In selecting the models, we focus on the following key aspects: 1) model diversity, 2) Open-Source vs. Closed-Source models, and 3) model parameter scaling law. 

\textbf{LLMs.}
To verify the critical role of images in answering questions, we specifically design an experiment in which some LLMs participate in the task without any image input. The purpose of this experiment is to assess whether these models can accurately understand the questions and make correct choices in the absence of image information, thereby further demonstrating the importance of images in the comprehension and problem-solving process. We select DeepSeek-67B, LLaMA-3-8B, and Qwen2-7b as the LLMs used in this experiment.

\textbf{Evaluation.}
We use accuracy as the primary evaluation metric, multi-choice format questions and answer extraction method, which are widely used in previous benchmarks such as Helleswag \citep{zellers2019hellaswag}, MMMU \citep{yue2023mmmu}, CMMMU \citep{zhang2024cmmmu}, MMLU \citep{li2024cmmlu} and so on. Since CII-Bench is entirely composed of multiple-choice questions, the evaluation process only requires extracting the selected option from the model's response, which simplifies the complexity of rule design. It is important to note that when models use chain-of-thought (CoT) prompts, the responses may include intermediate steps. Therefore, the evaluation rules must be sufficiently robust, or the model’s output must follow a fixed format. If the selected option cannot be extracted from the model's response, the model is considered to have answered the question incorrectly. For the detailed statistics of the model output, please see Appendix~\ref{sec:appendix-detail-results}. 
For reference, we also select three Chinese PhD students to evaluate human performance on CII-Bench.

\subsection{Main Results}
In this section, we conduct a comprehensive comparison of the performance of various MLLMs, LLMs, and humans on CII-Bench. Detailed results across different domains and emotional dimensions are presented in Table \ref{tab:overall_results}, while different image types, difficulty levels, and rhetoric can be found in Appendix~\ref{sec:appendix-other-results}. 
The main experimental results and findings are summarized as follows:

\begin{table*}[!t]
\centering
\scalebox{0.73}{
\begin{tabular}{lc|cccccc|ccc}
\toprule
 \textbf{Model}& \textbf{Overall} & \textbf{Life} & 
\textbf{Art} & \textbf{Society} & \textbf{Politics} & \textbf{Env.} & \textbf{CTC} & \textbf{Positive} & \textbf{Negative} & \textbf{Neutral} \\
 {} & (800) & (216) & (123) & (157) & (21) & (51) & (130) & (220) & (247) & (231) \\
\midrule
\multicolumn{11}{c}{\textit{Open-source Models}} \\ 
\midrule 
Qwen-VL-Chat & 34.3 & 27.9 & 34.7 & 32.5 & 45.8 & 55.2 & 36.5 & 34.0 & 35.1 & 33.6 \\
idefics2-8b & 36.3 & 25.0 & 46.3 & 38.1 & 41.7 & 56.9 & 32.9 & 32.8 & 39.1 & 36.4 \\
MiniCPM-Llama3-2.5 & 40.4 & 36.3 & 45.6 & 37.1 & 50.0 & 51.7 & 40.2 & 43.2 & 37.0 & 41.3 \\
CogVLM2-Llama3-Chinese-Chat & 43.4 & 37.1 & 48.3 & 42.3 & 54.2 & 63.8 & 40.2 & 40.3 & 45.7 & 43.8\\
MiniCPM-v2.6 & 45.0 & 37.5 & 47.6 & 49.5 & 58.3 & 55.2 & 42.3 & 45.6 & 44.6 & 44.9 \\
LLaVA-1.6-34B & 46.0 & 40.8 & \underline{55.1} & 42.8 & 45.8 & 62.1 & 43.1 & 44.4 & 48.2 & 45.2 \\
LLaVA-1.6-72B & 48.0 & 43.8 & 48.3 & 49.5 & \underline{70.8} & 60.3 & 43.8 & 41.5 & 52.5 & 49.2 \\
Qwen2-VL-7B & 49.6 & 42.5 & 51.7 & 54.1 & 62.5 & 65.5 & 44.5 & 50.2 & 47.5 & 51.2 \\
GLM-4V-9b & 50.3 & 46.7 & 48.3 & 53.6 & 54.2 & 62.1 & 48.2 & 51.9 & 52.9 & 46.3 \\
InternVL2-Llama3-76B & 52.9 & 50.8 & 53.7 & 51.0 & 58.3 & 67.2 & 51.1 & \underline{54.8} & 51.8 & 52.3 \\
InternVL2-8B & 53.1 & 49.2 & 53.1 & 55.7 & 62.5 & 63.8 & 50.4 & 50.6 & 53.3 & 55.1 \\
InternVL2-40B & \underline{57.9} & \underline{55.8} & \underline{55.1} & \underline{61.9} & 62.5 & \underline{70.7} & \underline{52.6} & 54.4 & \underline{58.0} & \underline{60.8} \\
Qwen2-VL-72B & \textbf{64.4} & \textbf{61.7} & \textbf{61.2} & \textbf{68.0} & \textbf{79.2} & \textbf{75.9} & \textbf{59.9} & \textbf{62.7} & \textbf{63.8} & \textbf{66.4} \\
\midrule
\multicolumn{11}{c}{\textit{Closed-source Models}} \\ 
\midrule
GPT-4o & 54.1 & 54.1 & 55.8 & 52.1 & 50.0 & 63.8 & 51.8 & 51.9 & 56.2 & 54.1 \\
Claude-3.5-Sonnet & 54.1 & 52.1 & \underline{61.9} & 52.6 & 62.5 & 46.6 & \underline{53.3} & 52.7 & 56.5 & 53.0 \\
Qwen-VL-MAX & 56.9 & 53.3 & 59.2 & 58.8 & 62.5 & \underline{67.2} & 52.6 & 53.9 & 58.3 & 58.0\\ 
Gemini-1.5 Pro & \underline{60.1} & \textbf{60.0} & \textbf{63.3} & \underline{62.4} & \textbf{70.8} & 62.1 & 51.1 & \underline{54.8} & \textbf{65.6} & \textbf{59.4} \\
GLM-4V & \textbf{60.9} & \underline{55.0} & 59.9 & \textbf{66.5} & \underline{66.7} & \textbf{79.3} & \textbf{55.5} & \textbf{58.5} & \underline{64.5} & \textbf{59.4} \\
\midrule
\multicolumn{11}{c}{\textit{Text-Only Models}} \\ 
\midrule
Llama-3-8B-Instruct & 21.7 & 22.2 & 26.9 & 18.6 & \underline{25.0} & 27.8 & \underline{20.4} & 21.2 & \underline{24.4} & 19.5 \\
DeepSeek-67B-Chat & \underline{27.1} & \underline{26.6} & \underline{32.7} & \textbf{30.9} & 20.0 & \underline{35.2} & 18.2 & \underline{25.7} & 22.2 & \underline{33.2} \\
Qwen2-7B-Instruct & \textbf{32.5} & \textbf{33.2} & \textbf{34.6} & \textbf{30.9} & \textbf{35.0} & \textbf{40.7} & \textbf{28.5} & \textbf{33.6} & \textbf{30.4} & \textbf{33.6} \\
\midrule
\multicolumn{11}{c}{\textit{Humans}} \\ 
\midrule
Human\_avg & 78.2 & 81.0 & 67.7 & 82.7 & 87.7 & 84.0 & 65.9 & 77.9& 75.2 & 81.6  \\ 
Human\_best & \textbf{81.0} & \textbf{83.2} & \textbf{73.6} & \textbf{87.2} & \textbf{89.5} & \textbf{86.0} & \textbf{66.7} & \textbf{78.2} & \textbf{78.8} & \textbf{83.3} \\ 
\bottomrule
\end{tabular}
}
\caption{Overall results of different MLLMs, LLMs and humans on different domains and emotions. The best-performing model in each category is \textbf{in-bold}, and the second best is \underline{underlined}.}
\label{tab:overall_results}
\end{table*}

\subsubsection{Natural Challenges of CII-Bench}

This benchmark presents a significant challenge for current models. Notably, despite GPT-4o being an advanced model, its accuracy is only 54.1\%, indicating substantial room for improvement. This reflects the rigorous and demanding nature of the benchmark. Further analysis reveals that most models perform worst in the domain of Chinese traditional culture, highlighting a significant deficiency in their understanding of Chinese cultural nuances. It is also noteworthy that human performance in this domain is not ideal, as questions related to Chinese traditional culture often require deep cultural knowledge. The lack of this knowledge base poses difficulties for both models and humans when dealing with Chinese cultural content.
In addition, text-only models like DeepSeek-67B-Chat only get 27.1\% accuracy, which shows that most of the questions in CII-Bench require image information to be answered correctly, proving that CII-Bench is highly dependent on visual content \citep{chen2024rightwayevaluatinglarge}.

\subsubsection{Gap between Humans and MLLMs}

The results indicate a significant gap between human performance and multimodal large models (MLLMs) on CII-Bench. Human participants achieved an average accuracy of 78.2\%, with the highest accuracy reaching 81.0\%. In contrast, the best-performing closed-source model, GLM-4V, achieved an accuracy of 60.9\%, while the top open-source model, Qwen2-VL-72B, scored 64.4\%. These findings highlight the substantial disparity between human abilities and even the most advanced models in understanding image implications. The highest accuracy achieved by the models is considerably lower than the average human score, indicating that multimodal large models still face significant challenges in this domain.

\subsubsection{Model Performance across Different Domains and Emotions} 

In terms of domain performance, our results in Table \ref{tab:overall_results} indicate that the models generally perform better in the Environment and Politics domains, achieving higher accuracy. Conversely, the accuracy is lower in the Life and Society domains, proving that everyday metaphors are generally more difficult in the Chinese context.
The lowest score for the Chinese Traditional Culture and Art domains, which shows that while the models generalize well in common domains, they struggle with the more abstract and logically demanding information found in Chinese Traditional Culture and Art.

From an emotional perspective, the models tend to exhibit higher accuracy when the image implications convey negative emotions, while accuracy is the lowest for images with positive emotions. This discrepancy highlights that the models' preferences do not align with those of humans, as humans are significantly more sensitive to positive implications. 
The performance of the model is opposite to the conclusion shown in II-Bench \citep{liu2024iibenchimageimplicationunderstanding}, reflecting the obvious difference in emotional expression in the Chinese and English contexts.

\subsubsection{Analysis on different prompt skills}

\begin{table*}[!t]
\centering
\scalebox{0.85}{
\setlength{\tabcolsep}{13.3pt}
\begin{tabular}{lccccc}
\toprule
\textbf{Model} & \textbf{None} & \textbf{CoT} & 
\textbf{Domain} & \textbf{Emotion} & \textbf{Rhetoric} \\
\midrule
\multicolumn{6}{c}{\textit{Open-source Models}} \\ 
\midrule
Qwen-VL-Chat & 34.3 & 34.0 & 32.1 & 35.0 & 33.4 \\
idefics2-8b & 36.3 & 33.3 & 37.5 & 38.6 & 37.4 \\
MiniCPM-Llama3-2.5 & 40.4 & 35.8 & 41.1 & 39.0 & 34.8 \\
CogVLM2-Llama3-Chinese-Chat & 43.4 & 42.6 & 43.5 & 44.0 & 43.4 \\
MiniCPM-v2.6 & 45.0 & 38.9 & 44.4 & 45.4 & 45.4 \\
LLaVA-1.6-34B & 46.0 & 44.5 & 46.4 & 47.1 & 45.4 \\
LLaVA-1.6-72B & 48.0 & 45.3 & 47.3 & 48.6 & 45.4 \\
Qwen2-VL-7B & 49.6 & 50.0 & 51.0 & 50.8 & 49.3 \\
GLM-4V-9b & 50.3 & 49.1 & 49.9 & 51.1 & 49.5 \\
InternVL2-Llama3-76B & 52.9 & 52.6 & 54.1 & 52.8 & 53.5 \\
InternVL2-8B & 53.1 & 47.9 & 53.5 & 56.3 & 53.8 \\
InternVL2-40B & \underline{57.9} & \underline{57.6} & \underline{57.1} & \underline{60.0} & \underline{57.9} \\
Qwen2-VL-72B & \textbf{64.4} & \textbf{62.1} & \textbf{66.0} & \textbf{64.3} & \textbf{63.0} \\
\midrule
\multicolumn{6}{c}{\textit{Closed-source Models}} \\ 
\midrule
GPT-4o & 54.1 & \textbf{54.9} & 55.4 & 54.9 & 51.9  \\
Claude-3.5-Sonnet & 54.1 & 51.6 & 56.4 & 53.5 & 54.9  \\
Qwen-VL-MAX & 56.9 & 54.0 & \underline{59.1} & \underline{59.9} & 54.8\\ 
Gemini-1.5 Pro  & \underline{60.1} & \underline{54.1} & 59.0 & 57.9 & \underline{55.6}  \\
GLM-4V & \textbf{60.9} & 48.8 & \textbf{60.4} & \textbf{60.6} & \textbf{58.8} \\
\bottomrule
\end{tabular}%
}
\caption{Overall results of different prompts on {CII-Bench}. The label (\textit{Emotion, Domain, Rhetoric}) means providing corresponding information for the images in the prompt. The best-performing model in each category is \textbf{in-bold}, and the second best is \underline{underlined}.}
\label{tab:prompt_results}
\vspace{-1em}
\end{table*}

\paragraph{Analysis of Chain-of-Thought (CoT).}

In Table \ref{tab:prompt_results}, we evaluate the impact of Chain-of-Thought (CoT) prompting on model performance. The results indicate that CoT does not significantly improve the accuracy of the models. In some cases, particularly with smaller open-source models, the accuracy even declined when CoT was used. For example, MiniCPM-v2.6 scores 45.0\% without CoT, but this drops to 38.9\% with CoT; similarly, LLaVA-1.6-72B scores decrease from 48.0\% to 45.3\%. 

Upon analyzing the models' responses, we find that those models showing a decrease in accuracy with CoT often suffer from overinterpretation, where questions that were initially answered correctly are misinterpreted after CoT is applied. Additionally, for questions that were originally answered incorrectly, CoT does not lead to significant improvements and sometimes even causes confusion, such as selecting multiple options. However, for most models, the probability of failing to extract an answer option from the response decreases after using CoT, which explains why some models show improved accuracy with CoT.

\paragraph{Analysis of Different Types and Domains.}

To evaluate the impact of different label information on model accuracy, we conduct an ablation study by providing relevant label information (such as emotion, domain, and rhetoric) in the prompts. The results in Table \ref{tab:prompt_results} show that emotion labels significantly improve model accuracy, followed by domain and rhetoric labels, both of which exhibit similar effectiveness.

This result aligns with human intuition. The answer options typically include negative, positive, and neutral choices. When the model receives emotional information, it can eliminate some irrelevant options, naturally leading to higher accuracy. In contrast, domain and rhetoric information generally do not effectively help the model eliminate options, resulting in more limited improvements. Additionally, from a model training perspective, models tend to have a more mature understanding of emotions, while specific nouns in rhetoric and domain labels are often custom-defined. During pre-training, the model may not have encountered a large number of descriptions for such specific nouns, making these labels less helpful in improving accuracy.

\paragraph{Analysis of Few-shot Examples.}

The results in Table~\ref{tab:few-shot_results} indicate that few-shot examples do not improve the models' accuracy. Specifically, performance declines as the number of examples increases. 
This decline can be attributed to the models' inferior capabilities in handling multiple images compared to single images, leading to a decrease in accuracy with a higher number of shots.
Furthermore, as the number of shots increases, the input length also extends, and the models' ability to process long texts is inadequate, resulting in suboptimal performance with long contexts.

\begin{table*}[htp]
\centering
\scalebox{0.85}{
\setlength{\tabcolsep}{13.3pt}
\begin{tabular}{lcccc}
\toprule
\textbf{Model} & \textbf{None} & \textbf{1-shot} & 
\textbf{2-shot} & \textbf{3-shot} \\
\midrule
Qwen2-VL-7B & 49.6 & 44.1 & 39.3 & 37.5 \\
GPT-4o & 54.1 & 51.8 & 49.5 & 49.1 \\
Claude-3.5-Sonnet & 54.1 & 55.4 & 55.3 & 55.4 \\
InternVL2-40B & 57.9  & 53.0 & 47.1 & 41.9 \\
Gemini-1.5 Pro & 60.1 & 57.4 & 55.8 & 55.4 \\
\bottomrule
\end{tabular}
}
\caption{Few-shot results of different models on the {CII-Bench}.}
\label{tab:few-shot_results}
\vspace{-1em}
\end{table*}

\subsection{Evaluation of Chinese Traditional Culture}

The Chinese traditional culture category is a distinctive feature of the CII-Bench dataset, where MLLMs consistently score the lowest. Therefore, we need a deeper evaluation of this field to analyze the extent to which MLLM understands Chinese traditional culture. 
We chose to deeply analyze MLLM's understanding of Chinese traditional culture by evaluating Chinese traditional paintings.

\subsubsection{Evaluation Metric}

Chinese traditional painting, a cornerstone of Chinese traditional culture, encompasses a rich tapestry of styles and techniques developed over millennia. These paintings are typically categorized based on their subject matter (e.g., landscape paintings, flower-and-bird paintings, figure paintings, and New Year paintings) or their stylistic and skill (e.g., court paintings, meticulous brush paintings, freehand brush paintings, and color-and-ink paintings). Each category embodies unique characteristics that reflect China's artistic evolution and philosophical underpinnings.

\begin{figure}[htbp]
  \centering  
  \includegraphics[width=0.85\textwidth]{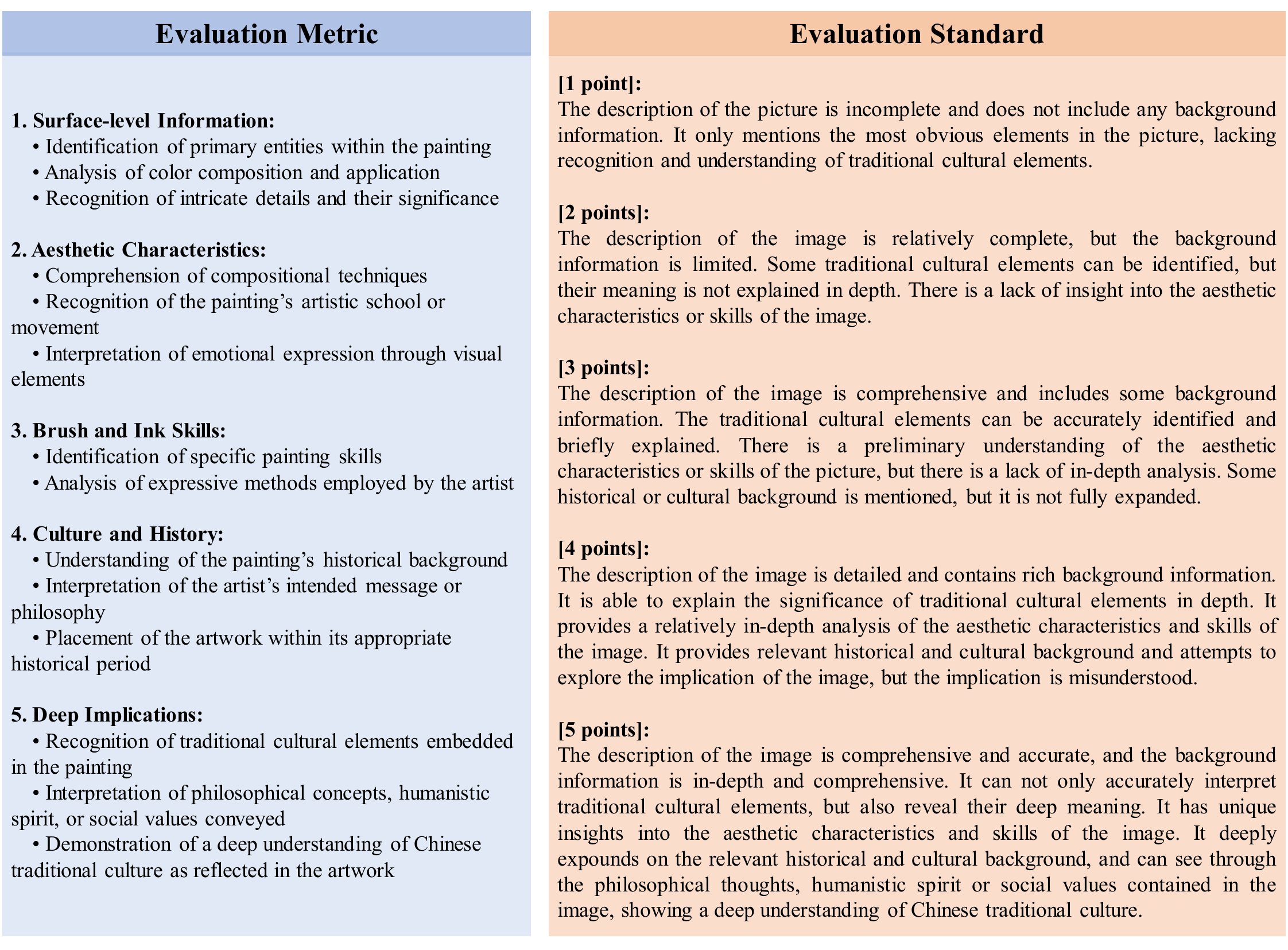}
  \caption{Evaluation metric and evaluation standard of Chinese traditional painting.}
  \label{figure:CTC_Evaluation}
\end{figure}

To comprehensively assess MLLMs' understanding of Chinese traditional paintings, we develop a multifaceted evaluation metric. This metric is designed to probe both the surface-level information readily apparent in the artwork and the deeper culture and history that informs its creation and interpretation. Our evaluation metric encompasses five key perspectives: \textbf{\textit{Surface-level Information}}, \textbf{\textit{Aesthetic Characteristics}}, \textbf{\textit{Brush and Ink Skills}}, \textbf{\textit{Culture and History}}, and \textbf{\textit{Deep Implications}}.  
For each perspective, we give its detailed description in Figure~\ref{figure:CTC_Evaluation}. 

\subsubsection{LLM-based Chinese Traditional Painting Automatic Evaluation}

To evaluate Chinese traditional painting comprehension in MLLMs, we develop an LLM-based evaluation standard based on evaluation metrics, as illustrated in Figure~\ref{figure:CTC_Evaluation}. Our experiment utilize the CTC domain data from CII-Bench, comprising 130 Chinese traditional paintings. We employ human-written descriptions and implication interpretations as ground truth. We choose GPT-4o to generate descriptions for these images, which are subsequently scored using GPT-4o and our evaluation standard. Please see the evaluation prompt in Appendix~\ref{sec:appendix-experiments_baseline}. To validate the model's scoring efficacy, we enlist three PhD students well-versed in Chinese metaphorical imagery to independently score the 130 paintings.

The model-human scoring consistency reached 98\%, affirming the method's validity for assessing Chinese traditional painting comprehension. Table~\ref{tab:CTC_results} presents the detailed model scores. Analysis of these results, in conjunction with our evaluation standard, reveals insights across three dimensions: overall performance, difficulty levels, and emotions. The overall score of 2.71 indicates that while MLLM is able to observe the surface-level information of paintings, it has a large gap with humans in deeply interpreting the complex cultural elements contained in Chinese traditional art. In terms of difficulty evaluation, the model is consistent with human cognition, while in terms of emotion, the model has a higher negative score, indicating that the model can identify negative implications in paintings, such as using the past to satirize the present, and not appreciating talents.

\begin{table*}[thbp]
\centering
\scalebox{0.85}{
\setlength{\tabcolsep}{9.5pt}
\begin{tabular}{lccccccc}
\toprule
\textbf{Model} & \textbf{Overall} & \textbf{Easy} & 
\textbf{Middle} & \textbf{Difficult} & \textbf{Positive} & \textbf{Negative} & \textbf{Neutral} \\
\midrule
GPT-4o & 2.71 & 3.0 & 3.2 & 2.35 & 2.63 & 3.0 & 2.82 \\
\bottomrule
\end{tabular}
}
\caption{Overall result of Chinese traditional painting.}
\label{tab:CTC_results}
\vspace{-1em}
\end{table*}

\subsection{Error Analysis}

\begin{wrapfigure}{r}{0.47\linewidth}
  \centering  
  \includegraphics[width=0.48\textwidth]{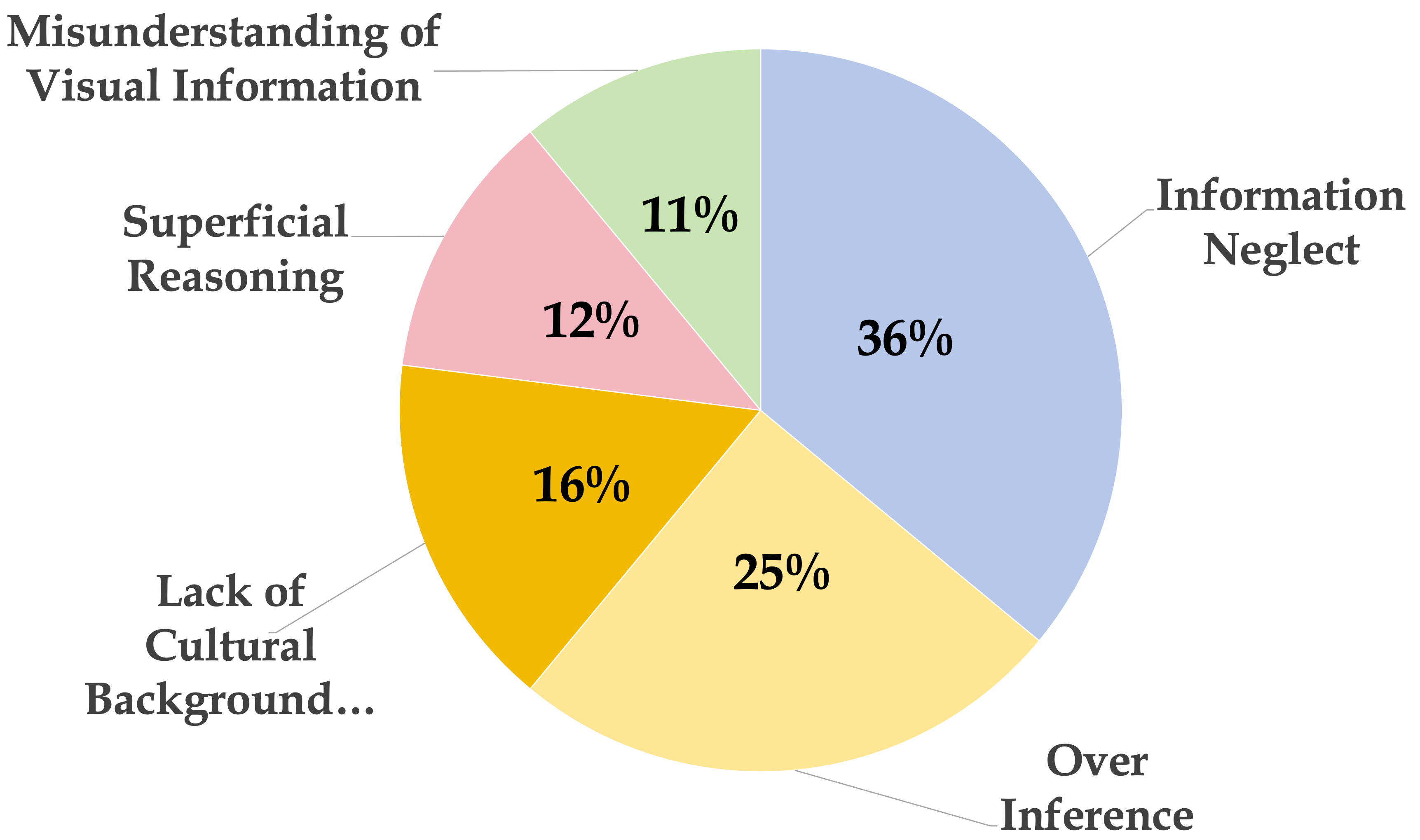}
  \caption{GPT-4o error responses distribution.}
  \label{figure:error}
\end{wrapfigure}

To conduct a comprehensive error analysis of GPT-4o's performance (under CoT setting) on CII-Bench, we randomly select a total of 100 erroneous samples from various domains, distributed according to their proportions in the dataset. These samples are subjected to in-depth analysis by expert annotators. As illustrated in Figure \ref{figure:error}, GPT-4o's errors can be categorized into the following types: Information Neglect, Misunderstanding of Visual Information, Over-Inference, Superficial Reasoning, and Lack of Cultural Background Knowledge. 
For detailed analysis of cases, please see the Appendix~\ref{sec:appendix-case study}.

\textbf{Information Neglect (36\%):}

Complex images contain both visual and textual elements. Sole reliance on visual information makes accurate interpretation challenging due to diversity in meaning. Incorporating textual information clarifies the author's emotional intent, aiding accurate interpretation. Unfortunately, GPT-4o frequently overlooks key visual (13\%) and textual (23\%) information. When directly asked about these elements, we find that GPT-4o can often answer correctly, indicating two main issues: 1) Insufficient image recognition abilities, and 2) Significant shortcomings in multimodal fusion, leading to underutilization of acquired information.

\textbf{Over-Inference (25\%):}

During answer construction, distractors are included at surface and deep levels. GPT-4o often selects more exaggerated, deep-level incorrect options, ignoring narrower but correct ones, especially in Chinese memes. This suggests that GPT-4o has a preference for selecting abstract options.

\textbf{Lack of Cultural Background Knowledge (16\%):}

CII-Bench requires a model's deep understanding of Chinese traditional culture. Lacking knowledge of traditional symbols, historical figures, and classical allusions, GPT-4o struggles with interpreting deeper implications within images. Despite reasonable Chinese language handling, the model's cultural deficiency affects its reasoning and performance.

\textbf{Superficial Reasoning (12\%):}

Understanding extended meanings within images is crucial. However, GPT-4o often only focus on surface-level elements, neglecting the deep implications and deeper cultural connotations behind them. This superficial reasoning hinders the model from fully grasping profound messages that the artist or designer intends to convey.

\textbf{Misunderstanding of Visual Information (11\%):}

Accurate identification of visual information is vital. We find that GPT-4o sometimes misidentifies visual elements within images, particularly when dealing with abstract images. The abstract nature of such images often stems from the inclusion of exaggerated imaginative elements, sometimes even defying physical laws. Therefore, correctly identifying these abstract elements requires the model to have a deep understanding of the essence of objects, a capability that current models clearly do not yet possess.

\section{Discussion}

\subsection{Interpretability Analysis of Chinese Image Implications}
The essence of Chinese image implications is deeply rooted in deep cultural heritage and complex contextual associations, which enables them to convey profound messages through nuanced expressions. For example, in traditional Chinese art forms such as landscape and New Year paintings, the imagery transcends mere depiction of nature or daily occurrences. Instead, it embodies emotions, philosophical insights, and societal norms through metaphorical and highly symbolic expressions. These symbols, like the pine tree, plum blossom, and crane, are not superficial meaning but are steeped in centuries of cultural tradition, representing resilience, purity, and longevity.

However, deciphering these complex messages can be challenging, particularly for those unfamiliar with the cultural and historical narratives tied to these symbols. This contrasts with English image implications, which often convey messages through more straightforward and explicit symbolism. As a result, the interpretability of Chinese image implications depends to some extent on reconstructing and resonating with the cultural context, which is what makes them unique: their meaning is not only visual but also culturally resonant, bridging across time and space.

Moreover, the interpretability of Chinese image implications has new changed in the modern era. Globalization and the surge of internet culture have intertwined foreign elements with traditional Chinese culture, birthing new symbols and implications. This intersection introduces additional layers of meaning, complicating the interpretation of traditional symbols. 

\subsection{Why Choose Chinese Traditional Paintings to Evaluate Chinese Traditional Culture?}

The imagery associated with Chinese traditional culture often embodies complex implications, encompassing customs, historical anecdotes, and legendary tales, making direct evaluation particularly challenging.
Chinese traditional painting, intrinsically intertwined with Chinese traditional culture, offers a viable proxy for this assessment. The unique value of Chinese traditional painting lies in its embodiment of Chinese cultural connotations, aesthetic implications, and distinctive artistic expression. The core philosophical concepts of Confucianism, Taoism, and Buddhism, along with their humanistic essence, have consistently permeated the entire trajectory of Chinese painting history.
Consequently, we have chosen to evaluate MLLMs' comprehension of Chinese traditional culture through an in-depth analysis of their understanding of Chinese traditional paintings. 

\section{Conclusion}

The development of CII-Bench marks a significant step forward in evaluating the capabilities of multimodal large models (MLLMs) and brings us closer to achieving expert artificial general intelligence (AGI). This benchmark promotes a deeper exploration of the higher-order theory of mind in MLLMs. Experimental results indicate that current MLLMs still exhibit a significant gap compared to humans in understanding the implications of images within a Chinese context. We found that most MLLMs lack a deep knowledge base of Chinese traditional culture, leading to a superficial understanding of this cultural content. Finally, the experiments showed that incorporating image emotion hints into prompts often improves model performance, suggesting that models still struggle with emotional understanding, which in turn leads to misinterpretation of implications. 
We believe that CII-Bench will inspire the academic community to further develop the next generation of multimodal foundational models that move toward expert AGI.

\section*{Limitations}
\label{limit}
We acknowledge several limitations in our study. Although CII-Bench is comprehensive, subjective elements can result in varying interpretations, impacting result consistency. In addition, in order to ensure high quality and practicability, our benchmark is not particularly large. The evaluation metrics may not fully capture the advanced understanding and reasoning capabilities of AI systems. These limitations underscore the necessity for continuous refinement and expansion of our benchmarks. Future work will focus on developing and incorporating more stringent and objective test sets to enhance the reliability and validity of our benchmark.

\section*{Ethics Statement}
\label{ethics}
In developing CII-Bench, we strictly adhere to ethical guidelines and legal regulations, ensuring fairness, transparency, inclusivity and respect for all stakeholders. We stress the importance of safeguarding privacy and intellectual property rights, underscoring our commitment to responsible and lawful data management. We have taken steps to anonymize any personal data to protect privacy and and have made every effort to minimize harmful or biased content. However, we recognize that biases can inadvertently arise and some information may be potentially offensive. We are committed to continuous monitoring and improvement to mitigate such biases. Furthermore, we encourage users of our dataset to employ it responsibly and to consider the ethical implications of their work, particularly in applications that may impact individuals or communities.


\bibliography{iclr2025_conference}
\bibliographystyle{iclr2025_conference}

\clearpage
\newpage
\appendix

\section{Statistics of CII-Bench}
\label{sec:appendix-Data-inf}

\begin{table}[h!]
    \centering
    \footnotesize
    \begin{minipage}[t]{0.48\textwidth}
        \centering
        \begin{tabular}{@{}ll@{}}
            \toprule
            \multicolumn{2}{l}{\textbf{Statistics}} \\ \cmidrule(r){1-2}
            Total Questions             & 800\\
            Total Images                & 698  \\
            Dev : Validation : Test     & 15  : 20  : 765 \\
            Easy : Medium : Hard        & 305 : 282 : 111 \\ 
            \midrule
            
            Average Question Length        & 10.54  \\
            Average Option Length          & 28.31   \\
            Average Explanation Length     & 121.06  \\
            \midrule
            
            Metaphor                & 562 \\
            Exaggerate              & 121 \\
            Symbolism               & 236 \\
            Visual Dislocation      & 42  \\
            Antithesis              & 13  \\
            Analogy                 & 19  \\
            Personification         & 73 \\
            Contrast                & 87 \\ \bottomrule
        \end{tabular}
    \end{minipage}
    \hspace{1pt}
    \begin{minipage}[t]{0.48\textwidth}
        \centering
        \begin{tabular}{@{}ll@{}}
            \toprule
            \multicolumn{2}{l}{\textbf{Statistics}} \\ \cmidrule(r){1-2}
            Life        & 216 (30.95\%)   \\
            Art         & 123 (17.62\%)   \\
            Society     & 157 (22.49\%)\\
            Environment & 51 (7.31\%)   \\
            Politics    & 21 (3.01\%) \\ 
            Chinese Traditional Culture  & 130 (18.62\%) \\     
            \midrule
           
            Positive        & 220 (31.52\%)  \\
            Neutral         & 247 (35.39\%)  \\
            Negative        & 231 (33.09\%)  \\    
            \midrule
           
            Illustration       & 178 (25.50\%) \\
            Meme               & 145 (20.77\%) \\
            Poster             & 87 (12.46\%) \\
            Multi-panel Comic  & 34 (4.87\%)  \\
            Single-panel Comic & 143 (20.49\%) \\
            Painting           & 119 (17.05\%)  \\  \bottomrule
        \end{tabular}
    \end{minipage}
    \vspace{0.3cm}
    \caption{General statistics of {CII-Bench}.}
    \label{tab:dataset_statistics}
\end{table}

\begin{figure}[htbp]
  \centering
  \includegraphics[width=0.85\textwidth]{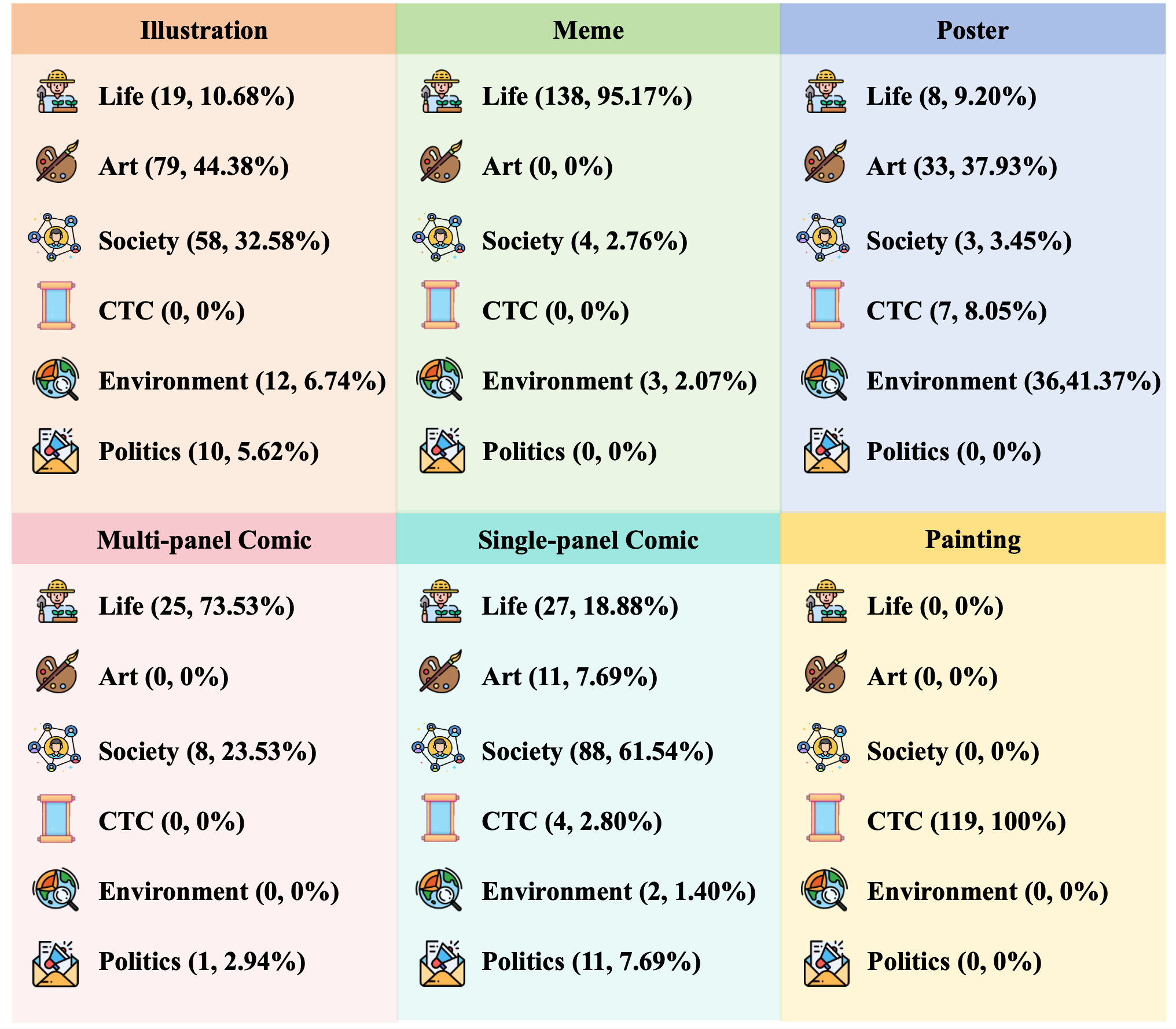}
  \caption{{CII-Bench} specific image type and domain statistics.}
  \label{figure:domain}
\end{figure}

\clearpage
\newpage

\section{CII-Bench Examples of English Version}
\label{sec:appendix-examples}

\begin{figure*}[htbp]
  \centering  
  \includegraphics[width=0.85\textwidth]{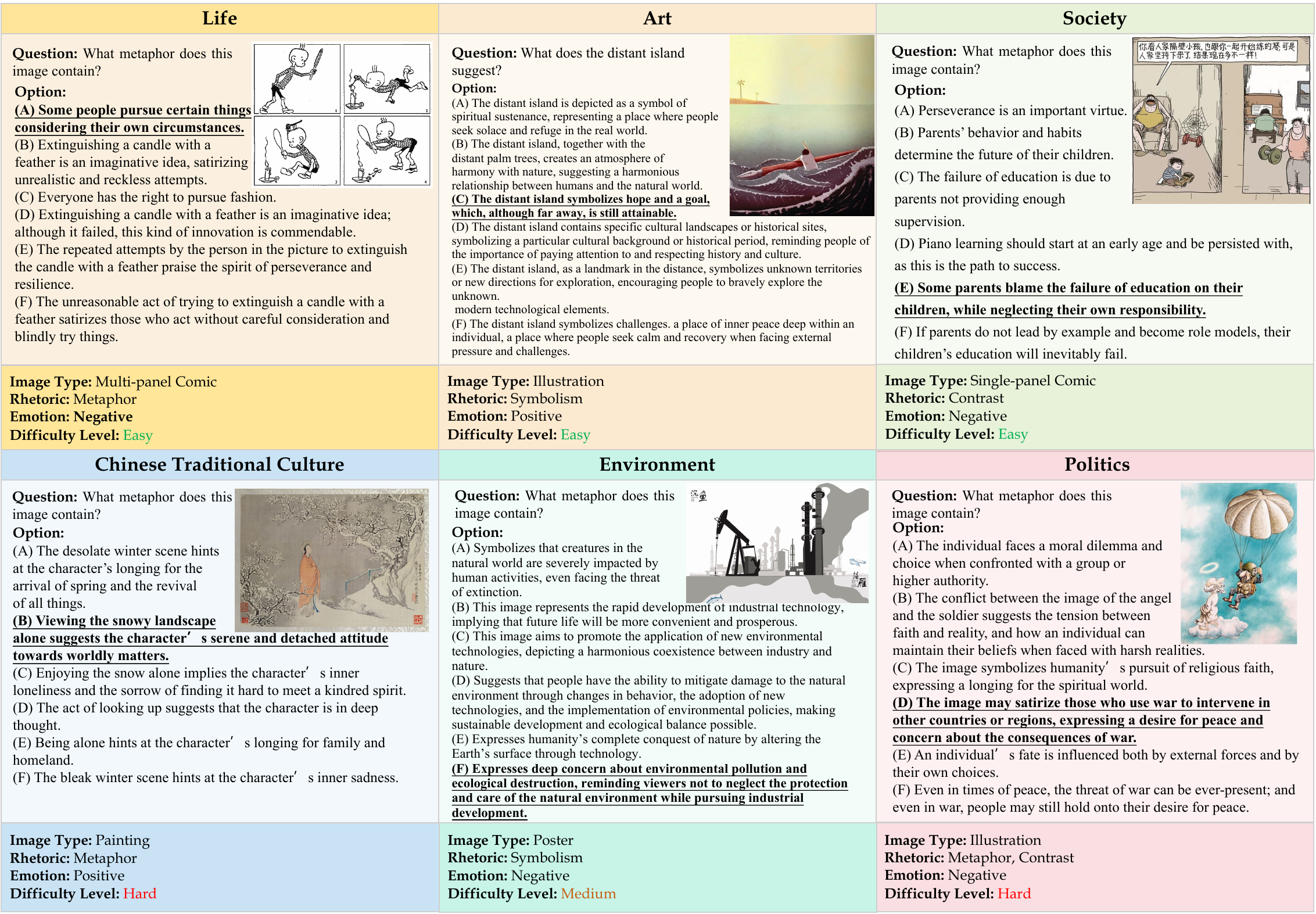}
  \caption{{CII-Bench} examples sampled from each domain. The pictures include life, art, society, Chinese traditional culture, environment and politics. Understanding these images and completing the corresponding questions require a certain level of comprehension.}
  \label{figure:sample_eng}
\end{figure*}

\clearpage
\newpage

\section{Data Annotation Protocol}
\label{sec:appendix-Data Annotation}
This document outlines a comprehensive protocol for annotating a dataset consisting of questions that explore the metaphorical implications of images.

\subsection{Data Collection}
Some websites from which we collect data are as follows: 
\begin{itemize}[left=5pt]
    \item \url{https://fabiaoqing.com/biaoqing/lists/page.html}
    \item \url{https://www.sohu.com/a/282205200_439969}
    \item \url{https://www.sohu.com/a/300233985_616741}
    \item \url{https://www.zcool.com.cn/u/746800}
    \item \url{https://www.shencaitang.com/news/1940}
    \item \url{https://www.dpm.org.cn/collection/paints.html}
    \item \url{https://www.zuomeme.com/wangyou/all}
\end{itemize}

\subsection{General Guidelines}
\textbf{General Principles:} 
\begin{itemize}[left=5pt]
    \item Annotations should be accurate and consistent.
    \item All questions, options and explanations should be written in Chinese.
    \item Any images without metaphorical implications should be discarded.
\end{itemize}

\textbf{Specific Instructions:}
\begin{itemize}[left=5pt]
    \item Each image needs to be categorized as one of the following image types: single-panel comic, multi-panel comic, poster, meme, illustration or painting.
    \item Each image needs to be categorized as one of the following difficulty levels from a human understanding perspective: easy, middle, or hard.
    \item Each image needs to be categorized as one of the following domains: life, art, society, politics, environment or Chinese traditional culture.
    \item Each image needs to be categorized as one of the following emotions: positive, neutral or negative.
    \item Each image needs to be categorized as one or more of the following rhetoric: metaphor, exaggerate, symbolism, contrast, visual dislocation, antithesis, analogy, personification or others.
    \item Each image needs a human explanation and implication description.
    \item Each image needs 1-3 questions about the fine-grained metaphorical implications of the image, each with one correct answer and five distractor options.
\end{itemize}

\subsection{Data Quality Assurance}
To further ensure the quality and reliability of the data, the annotated datasets were double-checked and cross-validated. Each question was manually validated by at least five annotators. Any inconsistencies or misinterpretations found were thoroughly examined and resolved by consensus of the annotation team, thus improving the reliability of the dataset while ensuring consistency of the annotations. In total, we conducted five rounds of data quality checks to ensure data quality and ultimately obtain CII-Bench.

\clearpage
\newpage

\subsection{Ethical Considerations}
\textbf{Copyright and Licensing.} It is essential to strictly follow all copyright and licensing regulations. Data from sources that do not permit copying or redistribution will be explicitly excluded.

\textbf{Data Privacy.} Adherence to privacy laws and ethical standards in data handling is crucial. Annotators must avoid collecting questions that contain any personal information.

\clearpage
\newpage

\section{Experiment Setup}
\label{sec:appendix-experiments_baseline}
\begin{figure*}[htbp]
    \centering
    \includegraphics[width=0.75\textwidth]{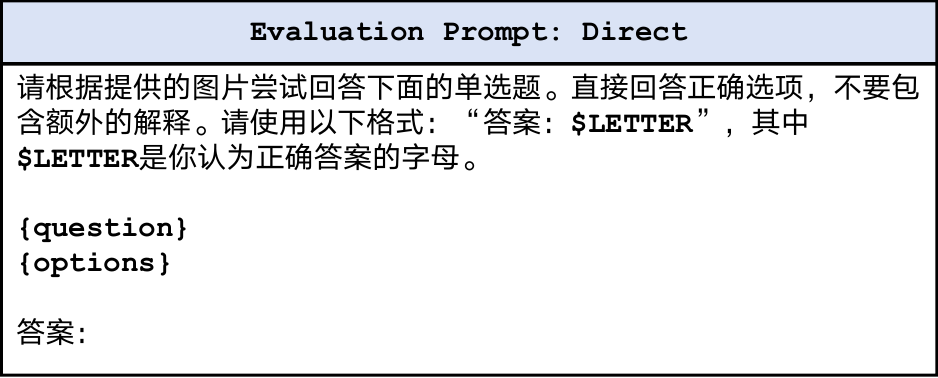}
    \caption{The prompt used in direct output setting.}
    \label{fig:prompt_direct}
\end{figure*}

\begin{figure*}[htbp]
    \centering
    \includegraphics[width=0.75\textwidth]{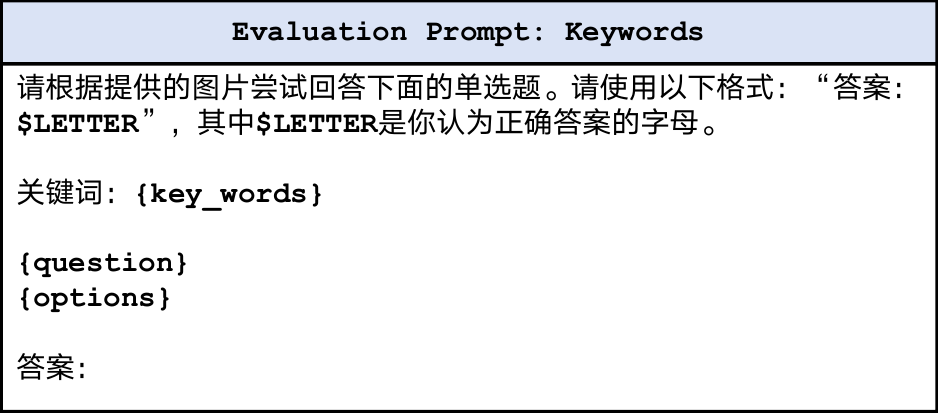}
    \caption{The prompt used in keyword setting.}
    \label{fig:prompt_keyword}
\end{figure*}

\begin{figure*}[htbp]
    \centering
    \includegraphics[width=0.75\textwidth]{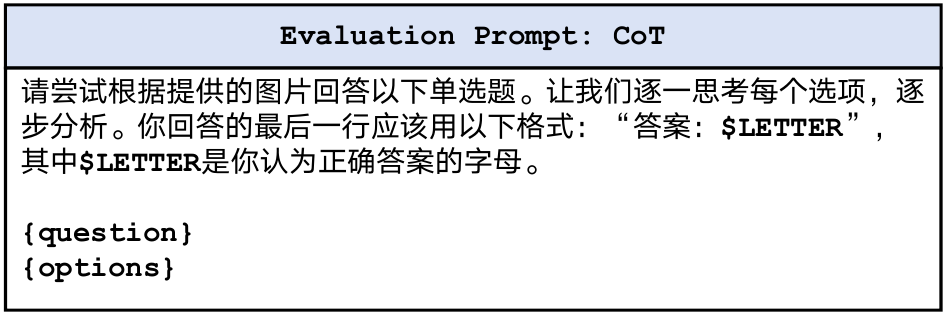}
    \caption{The prompt used in CoT setting.}
    \label{fig:prompt_cot}
\end{figure*}

\begin{figure*}[htbp]
    \centering
    \includegraphics[width=0.75\textwidth]{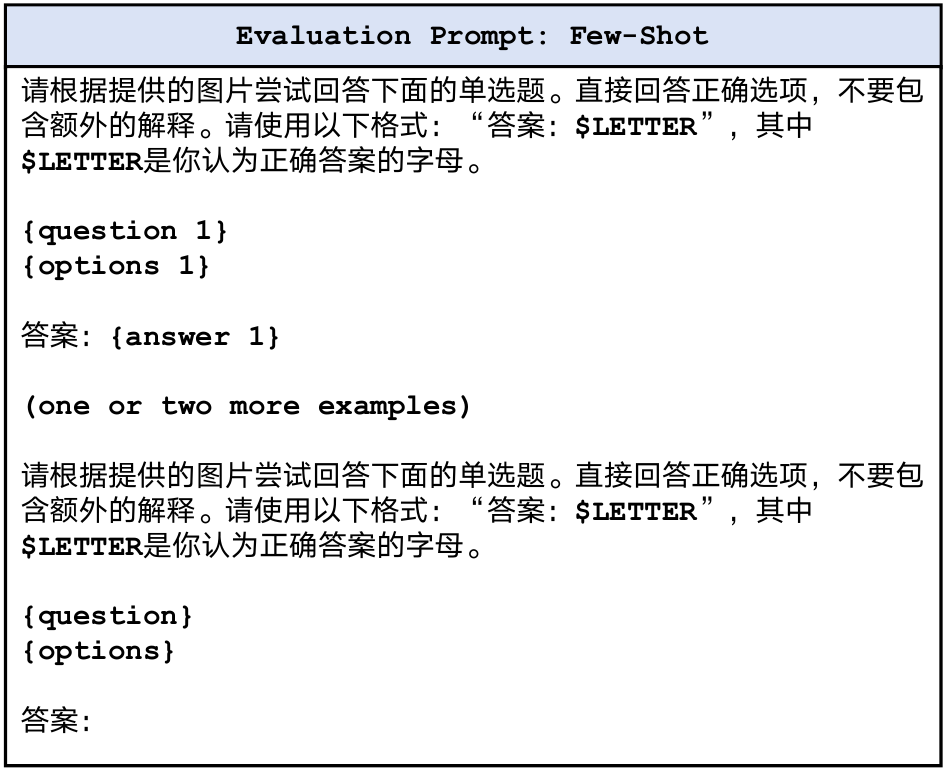}
    \caption{The prompt used in Few-Shot setting.}
    \label{fig:prompt_fewshot}
\end{figure*}

\begin{figure*}[htbp]
  \centering  
  \includegraphics[width=1\textwidth]{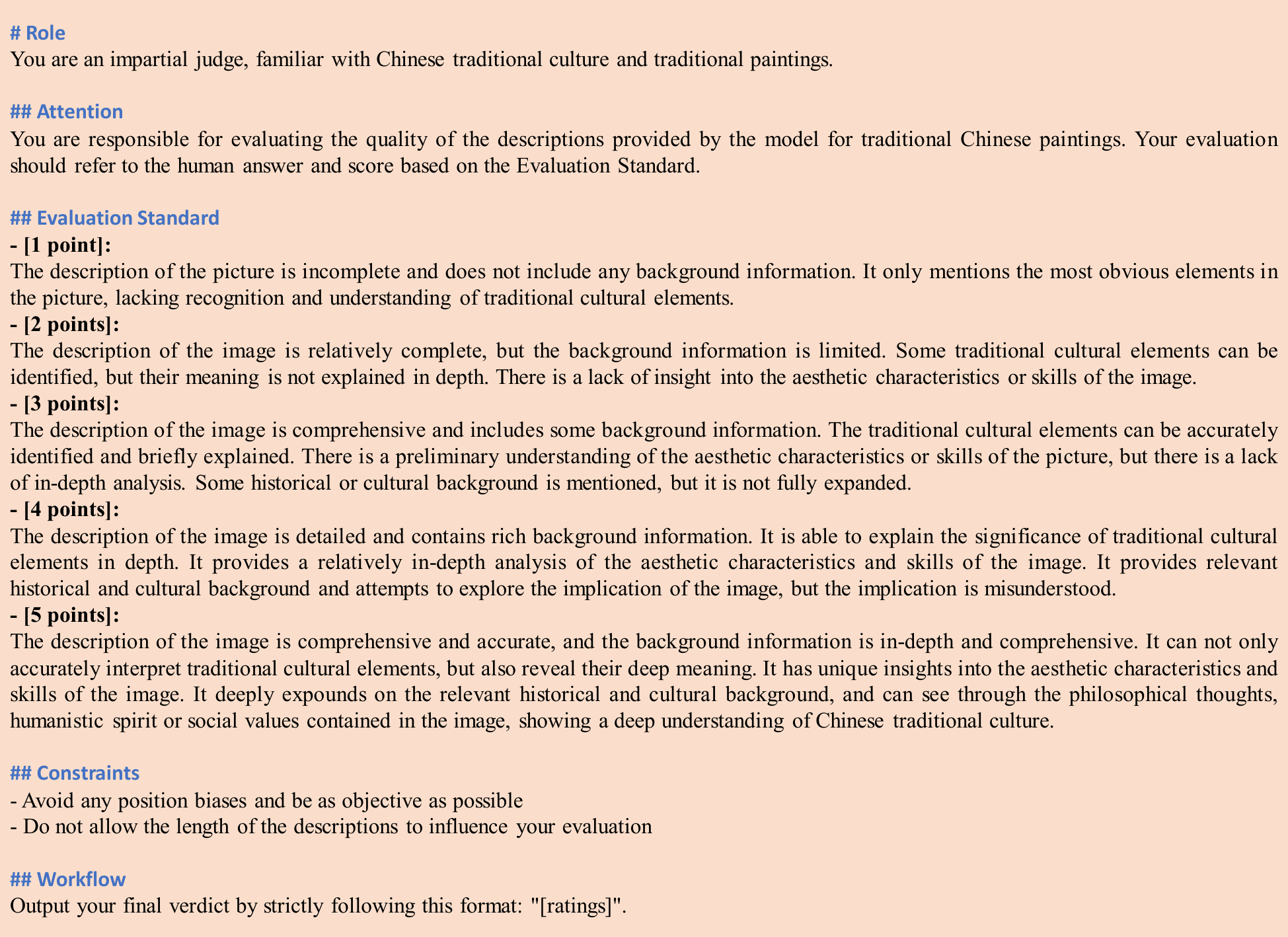}
  \caption{The prompt of Chinese traditional painting evaluation.}
  \label{figure:CTC_evaluation_prompt}
\end{figure*}

In experiments, we set the model temperature as 0, and all experiments are conducted on Nvidia A800 GPUs. The prompts of different settings are as follows Figure~\ref{fig:prompt_direct} to Figure~\ref{fig:prompt_fewshot}. Particularly, the evaluation prompt of Chinese traditional painting is Figure~\ref{figure:CTC_evaluation_prompt}.

\clearpage
\newpage
\input{appendix/result_appendix}

\clearpage
\newpage
\input{appendix/detail_appendix}

\clearpage
\newpage
\input{appendix/case_study}

\end{document}

%% file: math_commands.tex
\usepackage{amsmath,amsfonts,bm}

\def\eqref#1{equation~\ref{#1}}

\def\1{\bm{1}}

\DeclareMathAlphabet{\mathsfit}{\encodingdefault}{\sfdefault}{m}{sl}
\SetMathAlphabet{\mathsfit}{bold}{\encodingdefault}{\sfdefault}{bx}{n}



%% file: appendix/result_appendix.tex
\section{Results on Different Types, Difficulties and Rhetoric}
\label{sec:appendix-other-results}
In this section, we report the performance of different MLLMs and humans on different types of images, levels of difficulty, and rhetoric types.

\begin{table*}[htbp]
\centering
\scalebox{0.73}{
\begin{tabular}{lc|ccccccc}
\toprule
 \textbf{Model}& \textbf{Overall} & \textbf{Illus.} & \textbf{Paint.} & \textbf{Poster} & \textbf{Single-C.} & \textbf{Multi-C.} & \textbf{Meme} \\
\midrule
\multicolumn{8}{c}{\textit{Open-source Models}} \\ 
\midrule 
Qwen-VL-Chat & 34.3 & 33.5 & 36.8 & 45.1 & 35.2 & 23.7 & 27.5 \\
idefics2-8b & 36.3 & 44.0 & 32.8 & 45.1 & 35.2 & 23.7 & 24.8 \\
MiniCPM-Llama3-2.5 & 40.4 & 39.5 & 38.4 & 49.0 & 42.6 & 34.2 & 37.3 \\
CogVLM2-Llama3-Chinese-Chat & 43.4 & 45.0 & 39.2 & 52.9 & 45.5 & 23.7 & 39.2 \\
MiniCPM-v2.6 & 45.0 & 44.0 & 40.8 & 53.9 & 51.1 & 36.8 & 39.2 \\
LLaVA-1.6-34B & 46.0 & 50.0 & 44.0 & 48.0 & 47.7 & 29.0 & 42.5 \\
LLaVA-1.6-72B & 48.0 & 50.9 & 44.0 & 43.1 & 56.8 & 39.5 & 43.1 \\
Qwen2-VL-7B & 49.6 & 47.7 & 43.2 & \underline{0.8} & 58.0 & 31.6 & 46.4 \\
GLM-4V-9b & 50.3 & 46.8 & 47.2 & 55.9 & 59.7 & 42.1 & 47.1 \\
InternVL2-Llama3-76B & 52.9 & 48.2 & 50.4 & 59.8 & 62.5 & 39.5 & 49.7 \\
InternVL2-8B & 53.1 & 48.2 & 48.0 & 56.9 & 64.8 & \textbf{52.6} & 51.0 \\
InternVL2-40B & \underline{57.9} & \underline{53.7} & \underline{51.2} & 56.9 & \underline{68.2} & \underline{50.0} & \underline{59.5} \\
Qwen2-VL-72B & \textbf{64.4} & \textbf{61.5} & \textbf{59.2} & \textbf{68.6} & \textbf{70.5} & 47.4 & \textbf{67.3} \\
\midrule
\multicolumn{8}{c}{\textit{Closed-source Models}} \\ 
\midrule
GPT-4o & 54.1 & 54.1 & 50.4 & 56.9 & 54.6 & 47.4 & \underline{57.5} \\
Claude-3.5-Sonnet & 54.1 & 55.1 & \textbf{54.4} & 47.1 & 55.1 & \underline{50.0} & \underline{57.5} \\
Qwen-VL-MAX & 56.9 & 57.3 & 51.2 & \underline{60.8} & 62.5 & 39.5 & 56.2 \\ 
Gemini-1.5 Pro & \underline{60.1} & \textbf{64.7} & 50.4 & 52.0 & \underline{66.5} & \textbf{52.6} & \textbf{62.1} \\
GLM-4V & \textbf{60.9} & \underline{59.6} & \textbf{54.4} & \textbf{67.7} & \textbf{70.5} & 44.7 & \underline{57.5} \\
\midrule
\multicolumn{8}{c}{\textit{Humans}} \\ 
\midrule
Human\_avg & 78.2 & 71.5 & 65.6 & 75.2 & 79.8 & 74.5 & 83.6 \\ 
Human\_best & \textbf{81.0} & \textbf{76.9} & \textbf{66.1} & \textbf{78.6} & \textbf{81.7} & \textbf{78.4} & \textbf{85.0}\\ 
\bottomrule
\end{tabular}
}
\caption{Overall results of different MLLMs on different image types. The best-performing model in each category is \textbf{in-bold}, and the second best is \underline{underlined}. For brevity, Illus. refers to Illustration, Paint. refers to Painting, Single-C. refers to Single-panel Comic, Multi-C. refers to Multi-panel Comic.}
\label{tab:appendix-results-type}
\end{table*}

\begin{table*}[htbp]
\centering
\scalebox{0.73}{
\begin{tabular}{l|c|ccc}
\toprule
 \textbf{Model} & \textbf{Overall} & \textbf{Easy} & \textbf{Medium} & \textbf{Hard} \\
\midrule
\multicolumn{5}{c}{\textit{Open-source Models}} \\ 
\midrule 
Qwen-VL-Chat & 34.3 & 36.3 & 33.5 & 30.3 \\
idefics2-8b & 36.3 & 35.4 & 39.3 & 30.3 \\
MiniCPM-Llama3-2.5 & 40.4 & 43.1 & 39.3 & 35.3 \\
CogVLM2-Llama3-Chinese-Chat & 43.4 & 46.3 & 39.9 & 44.3 \\
MiniCPM-v2.6 & 45.0 & 47.1 & 44.2 & 41.0 \\
LLaVA-1.6-34B & 46.0 & 44.9 & 47.0 & 46.7 \\
LLaVA-1.6-72B & 48.0 & 50.0 & 47.0 & 45.1 \\
Qwen2-VL-7B & 49.6 & 52.6 & 47.9 & 45.9 \\
GLM-4V-9b & 50.3 & 52.6 & 49.1 & 46.7 \\
InternVL2-Llama3-76B & 52.9 & 57.4 & 49.7 & 48.4 \\
InternVL2-8B & 53.1 & 57.7 & 49.4 & 50.0 \\
InternVL2-40B & \underline{57.9} & \underline{62.3} & \underline{55.5} & \underline{51.6} \\
Qwen2-VL-72B & \textbf{64.4} & \textbf{68.9} & \textbf{63.1} & \textbf{54.9} \\
\midrule
\multicolumn{5}{c}{\textit{Closed-source Models}} \\ 
\midrule
GPT-4o & 54.1 & 56.0 & 54.9 & 46.7 \\
Claude-3.5-Sonnet & 54.1 & 55.1 & 52.4 & \underline{55.7} \\
Qwen-VL-MAX & 56.9 & 57.4 & 56.7 & \underline{55.7} \\
Gemini-1.5 Pro & \underline{60.1} & \underline{61.1} & \textbf{61.3} & 54.1 \\
GLM-4V & \textbf{60.9} & \textbf{62.9} & \underline{59.2} & \textbf{59.8} \\
\midrule
\multicolumn{5}{c}{\textit{Humans}} \\ 
\midrule
Human\_avg & 78.2 & 82.5 & 76.1 & 70.9 \\ 
Human\_best & \textbf{81.0} & \textbf{84.0} & \textbf{78.9} & \textbf{71.8} \\ 
\bottomrule
\end{tabular}
}
\caption{Overall results of different MLLMs on various difficulty levels. The best-performing model in each category is \textbf{in-bold}, and the second best is \underline{underlined}. The numbers in parentheses indicate the number of samples in each category.}
\label{tab:appendix-results-difficulty}
\end{table*}

\begin{table*}[htbp]
\centering
\scalebox{0.8}{
\begin{tabular}{lc|cccccccc}
\toprule
 \textbf{Model}& \textbf{Overall} & \textbf{Meta.} & \textbf{Exag.} & \textbf{Symb.} & \textbf{Contrast} & \textbf{VisD.} & \textbf{Pers.} & \textbf{Anal.} & \textbf{Anti.}  \\
\midrule
\multicolumn{10}{c}{\textit{Open-source Models}} \\ 
\midrule 
Qwen-VL-Chat & 34.3 & 31.8 & 38.9 & 38.4 & 41.0 & 37.0 & 34.2 & 28.6 & 30.8 \\
idefics2-8b & 36.3 & 35.2 & 32.6 & 35.6 & 41.9 & 30.4 & 26.6 & 23.8 & 38.5 \\
MiniCPM-Llama3-2.5 & 40.4 & 38.5 & 42.4 & 40.2 & 38.1 & 34.8 & 44.3 & 33.3 & 38.5\\
CogVLM2-Llama3-Chinese-Chat & 43.4 & 42.2 & 46.5 & 42.7 & 44.8 & 50.0 & 44.3 & 52.4 & 38.5  \\
MiniCPM-v2.6 & 45.0 & 41.7 & 48.6 & 43.4 & 41.0 & 45.7 & 45.6 & 38.1 & \textbf{53.9} \\
LLaVA-1.6-34B & 46.0 & 45.1 & 47.9 & 45.9 & 41.0 & 45.7 & 44.3 & 42.9 & 30.8 \\
LLaVA-1.6-72B & 48.0 & 46.1 & 54.2 & 48.0 & 49.5 & 47.8 & 46.8 & 47.6 & 38.5  \\
Qwen2-VL-7B & 49.6 & 47.6 & 52.1 & 48.4 & 49.5 & \underline{56.5} & 51.9 & 47.6 & \textbf{53.9}  \\
GLM-4V-9b & 50.3 & 48.7 & 56.3 & 51.3 & 52.4 & 50.0 & 50.6 & 57.1 & 30.8  \\
InternVL2-Llama3-76B & 52.9 & 51.5 & 59.7 & 51.3 & 51.4 & 52.2 & 55.7 & 52.4 & 46.2  \\
InternVL2-8B & 53.1 & 51.0 & 54.9 & 55.2 & 47.6 & 54.4 & 57.0 & 47.6 & 46.2  \\
InternVL2-40B & \underline{57.9} & \underline{55.8} & \underline{63.2} & \underline{56.6} & \underline{55.2} & 54.4 & \textbf{69.6} & \textbf{71.4} & 46.2  \\
Qwen2-VL-72B & \textbf{64.4} & \textbf{62.5} & \textbf{70.1} & \textbf{65.8} & \textbf{63.8} & \textbf{73.9} & \underline{67.1} & \underline{66.7} & \textbf{53.9} \\
\midrule
\multicolumn{10}{c}{\textit{Closed-source Models}} \\ 
\midrule
GPT-4o & 54.1 & 52.6 & 54.9 & 51.6 & 51.4 & \underline{60.9} & \underline{55.7} & 52.4 & 38.5 \\
Claude-3.5-Sonnet & 54.1 & 52.1 & 54.9 & 56.6 & 47.6 & 50.0 & 54.4 & 57.1 & 38.5  \\
Qwen-VL-MAX & 56.9 & 54.7 & 60.4 & 58.7 & 52.4 & 58.7 & \underline{55.7} & 57.1 & \underline{46.2}  \\ 
Gemini-1.5 Pro & \underline{60.1} & \underline{59.5} & \underline{64.6} & \underline{60.1} & \textbf{61.9} & 47.8 & \underline{55.7} & \textbf{81.0} & \textbf{53.9}  \\
GLM-4V & \textbf{60.9} & \textbf{60.2} & \textbf{65.3} & \textbf{63.4} & \underline{57.1} & \textbf{65.2} & \textbf{60.8} & \underline{66.7} & \underline{46.2}  \\
\midrule
\multicolumn{10}{c}{\textit{Humans}} \\ 
\midrule
Human\_avg & 78.2 & 76.0 & 82.8 & 74.1 & 70.4 & 73.9 & 72.9 & 90.0 & 52.8\\ 
Human\_best & \textbf{81.0} &\textbf{ 77.0} & \textbf{85.2} & \textbf{76.5} & \textbf{75.7} & \textbf{75.6} & \textbf{74.7} & \textbf{95.0} & \textbf{66.7}\\ 
\bottomrule
\end{tabular}
}
\caption{Overall results of different MLLMs and humans on different rhetoric. The best-performing model in each category is \textbf{in-bold}, and the second best is \underline{underlined}. For brevity, Meta. refers to Metaphor, Exag. refers to Exaggerate, Symb. refers to Symbolism, VisD. refers to Visual Dislocation, Anti. refers to Antithesis, Anal. refers to Analogy, Pers. refers to Personification }
\label{tab:appendix-results-rhetoric}
\end{table*}

%% file: appendix/detail_appendix.tex
\section{Additional Details of Results}
\label{sec:appendix-detail-results}

We do detailed statistics of the model output. The results are shown in Table \ref{tab:appendix-addtional-results-1}~to~\ref{tab:appendix-addtional-results-4}.
\textit{Miss} is mainly caused by two situations, one is that the model does not give an answer, and the other is the regex is not matched. The \textit{Miss} rate of most models is controlled below an acceptable ratio. In the \textit{CoT} setting, some models do not follow instructions well and do not provide the expected letters as answer, which cannot be matched and will be considered a \textit{Miss}.

\begin{table*}[!h]
\centering
\scalebox{0.75}{
\begin{tabular}{ccccccc}
    \toprule
    \textbf{Mode} & \textbf{Metric} & \textbf{InternVL2-40B} & \textbf{InternVL2-8B} & \textbf{InternVL2-Llama3-76B} & \textbf{MiniCPM-Llama3-2.5} & \textbf{MiniCPM-v2.6} \\
    \midrule
    \multirow{3}{*}{CoT} 
    & Acc & 57.6 & 47.9 & 52.6 & 35.8 & 39.3 \\
    & Error & 0.0 & 0.0 & 0.0 & 0.0 & 0.0 \\
    & Miss & 0.0 & 0.0 & 0.0 & 8.1 & 0.0 \\
    \midrule
    \multirow{3}{*}{Domain} 
    & Acc & 57.1 & 53.5 & 54.1 & 41.1 & 44.4 \\
    & Error & 0.0 & 0.0 & 0.0 & 0.0 & 0.0 \\
    & Miss & 0.0 & 0.0 & 0.0 & 5.9 & 0.0 \\
    \midrule
    \multirow{3}{*}{Emotion} 
    & Acc & 60.0 & 56.3 & 52.8 & 39.0 & 45.4 \\
    & Error & 0.0 & 0.0 & 0.0 & 0.0 & 0.0 \\
    & Miss & 0.0 & 0.0 & 0.0 & 8.4 & 0.0 \\
    \midrule
    \multirow{3}{*}{None} 
    & Acc & 57.9 & 53.1 & 52.9 & 40.4 & 45.0 \\
    & Error & 0.0 & 0.0 & 0.0 & 0.0 & 0.0 \\
    & Miss & 0.0 & 0.0 & 0.0 & 0.4 & 0.0 \\
    \midrule
    \multirow{3}{*}{Rhetoric} 
    & Acc & 57.9 & 53.8 & 53.5 & 34.8 & 45.4 \\
    & Error & 0.0 & 0.0 & 0.0 & 0.0 & 0.0 \\
    & Miss & 0.0 & 0.0 & 0.0 & 10.4 & 0.0 \\
    \bottomrule
\end{tabular}}
\caption{Accuracy, Error and Miss rate of different models under different settings.(1/4)}
\label{tab:appendix-addtional-results-1}
\end{table*}

\vspace{1cm}

\begin{table*}[!h]
\centering
\scalebox{0.8}{
\begin{tabular}{ccccccc}
    \toprule
    \textbf{Mode} & \textbf{Metric} & \textbf{Qwen-VL-Chat} & \textbf{Qwen2-VL-72B} & \textbf{Qwen2-VL-7B} & \textbf{CogVLM2-Llama3-Chinese-Chat} \\ 
    \midrule
    \multirow{3}{*}{CoT} & Acc & 34.0 & 62.1 & 50.0 & 43.0 \\
    & Error & 0.3 & 0.0 & 0.0 & 0.0  \\
    & Miss & 0.0 & 0.0 & 0.3 & 0.0  \\
    \midrule
    \multirow{3}{*}{Domain} & Acc & 32.1 & 66.0 & 51.0 & 43.5  \\
    & Error & 0.3 & 0.0 & 0.0 & 0.0  \\
    & Miss & 0.1 & 0.0 & 0.0 & 0.0  \\
    \midrule
    \multirow{3}{*}{Emotion} & Acc & 35.0 & 64.3 & 50.8 & 44.0  \\
    & Error & 0.1 & 0.0 & 0.0 & 0.0 \\
    & Miss & 0.5 & 0.0 & 0.0 & 0.0  \\
    \midrule
    \multirow{3}{*}{None} & Acc & 34.3 & 64.4 & 49.6 & 43.4  \\
    & Error & 0.5 & 0.0 & 0.0 & 0.0  \\
    & Miss & 0.4 & 0.0 & 0.0 & 0.0  \\
    \midrule
    \multirow{3}{*}{Rhetoric} & Acc & 33.4 & 63.0 & 49.3 & 43.4  \\
    & Error & 0.3 & 0.0 & 0.0 & 0.0  \\
    & Miss & 0.3 & 0.0 & 0.0 & 0.0  \\
    \bottomrule
\end{tabular}}
\caption{Accuracy, Error and Miss rate of different models under different settings.(2/4)}
\label{tab:appendix-addtional-results-2}
\end{table*}

\begin{table*}[!h]
\centering
\scalebox{0.85}{
\begin{tabular}{cccccc}
\toprule
\textbf{Mode} & \textbf{Metric} & \textbf{GLM-4V-9b} & \textbf{LLaVA-1.6-72B} & \textbf{LLaVA-1.6-34B} & \textbf{idefics2-8b} \\ 
    \midrule
    \multirow{3}{*}{CoT} & Acc & 49.1 & 45.3 & 44.5 & 33.3 \\
    & Error & 0.0 & 0.0 & 0.0 & 0.0 \\
    & Miss & 0.0 & 0.0 & 0.0 & 0.0 \\
    \midrule
    \multirow{3}{*}{Domain} & Acc & 49.9 & 47.3 & 46.4 & 37.5 \\
    & Error & 0.0 & 0.0 & 0.0 & 0.0 \\
    & Miss & 0.0 & 0.0 & 0.0 & 0.0 \\
    \midrule
    \multirow{3}{*}{Emotion} & Acc & 51.1 & 48.6 & 47.1 & 38.6 \\
    & Error & 0.0 & 0.0 & 0.0 & 0.0 \\
    & Miss & 0.0 & 0.0 & 0.0 & 0.1 \\
    \midrule
    \multirow{3}{*}{None} & Acc & 50.3 & 48.0 & 46.0 & 36.3 \\
    & Error & 0.0 & 0.0 & 0.0 & 0.0 \\
    & Miss & 0.0 & 0.0 & 0.0 & 0.0 \\
    \midrule
    \multirow{3}{*}{Rhetoric} & Acc & 49.5 & 45.4 & 45.4 & 37.4 \\
    & Error & 0.0 & 0.0 & 0.0 & 0.0 \\
    & Miss & 0.0 & 0.0 & 0.0 & 0.0 \\
    \bottomrule
\end{tabular}}
\caption{Accuracy, Error and Miss rate of different models under different settings.(3/4)}
\label{tab:appendix-addtional-results-3}
\end{table*}

\vspace{-1em}

\begin{table*}[!h]
\centering
\scalebox{0.85}{
\begin{tabular}{ccccccc}
\toprule
\textbf{Mode} & \textbf{Metric} & \textbf{Gemini-1.5 Pro} & \textbf{GLM-4V} & \textbf{GPT-4o} & \textbf{Claude-3-5-Sonnet} & \textbf{Qwen-VL-MAX} \\ 
    \midrule
    \multirow{3}{*}{CoT} & Acc & 54.1 & 49.9 & 54.9 & 51.6 & 54.8 \\
    & Error & 0.3 & 3.4 & 0.0 & 1.8 & 1.1 \\
    & Miss & 1.8 & 2.4 & 0.1 & 0.0 & 0.0 \\
    \midrule
    \multirow{3}{*}{Domain} & Acc & 59.0 & 60.4 & 55.4 & 56.4 & 59.1 \\
    & Error & 0.3 & 1.6 & 0.0 & 2.5 & 1.5 \\
    & Miss & 1.4 & 0.0 & 0.0 & 0.0 & 0.1 \\
    \midrule
    \multirow{3}{*}{Emotion} & Acc & 58.0 & 60.6 & 54.9 & 53.5 & 59.9 \\
    & Error & 0.3 & 3.4 & 0.0 & 2.5 & 1.1 \\
    & Miss & 1.8 & 0.0 & 0.1 & 0.0 & 0.0 \\
    \midrule
    \multirow{3}{*}{None} & Acc & 60.1 & 60.9 & 54.1 & 54.1 & 56.9 \\
    & Error & 0.3 & 0.0 & 0.0 & 3.3 & 1.9 \\
    & Miss & 0.1 & 0.0 & 0.0 & 0.9 & 0.0 \\
    \midrule
    \multirow{3}{*}{Rhetoric} & Acc & 55.6 & 58.8 & 51.9 & 54.9 & 54.8 \\
    & Error & 0.3 & 2.1 & 0.0 & 1.9 & 0.9 \\
    & Miss & 0.9 & 0.0 & 0.1 & 0.0 & 0.0 \\
    \bottomrule
\end{tabular}}
\caption{Accuracy, Error and Miss rate of different models under different settings.(4/4)}
\label{tab:appendix-addtional-results-4}
\end{table*}

%% file: appendix/case_study.tex
\section{case study}
\captionsetup[figure]{labelformat=simple, labelsep=colon, name=Figure}
\renewcommand{\thefigure}{G\arabic{figure}}
\setcounter{figure}{0}
\label{sec:appendix-case study}
The appendix is our sample analysis of GPT-4o, including an analysis of six error examples.

\hypertarget{listofcasestudyfigures}{}
\listofcasestudyfigures

\casestudyfigure{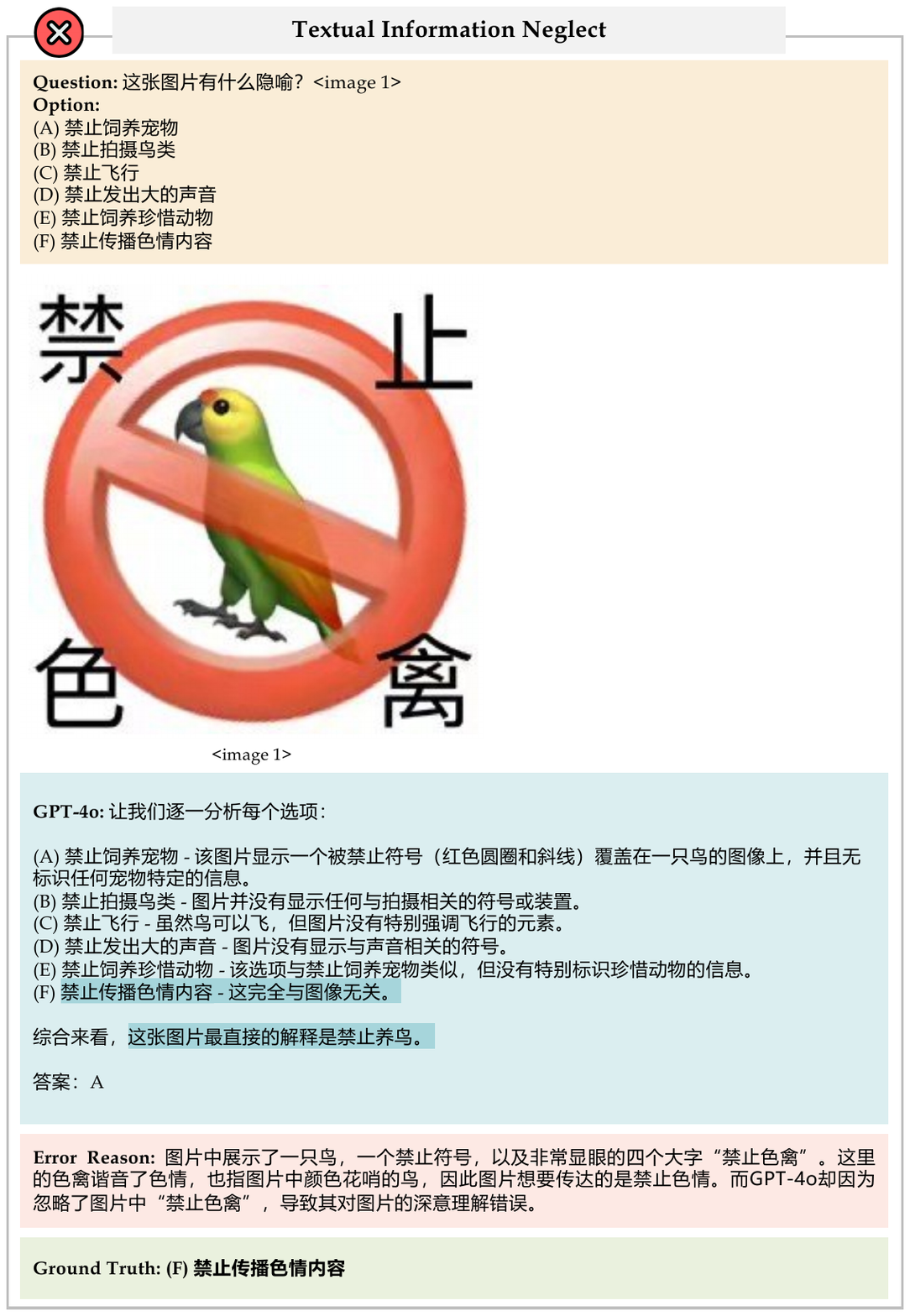}{Error 1: Textual Information Neglect}{A sample error case of \textit{Information Neglect}.}{fig:case_study_1}

\casestudyfigure{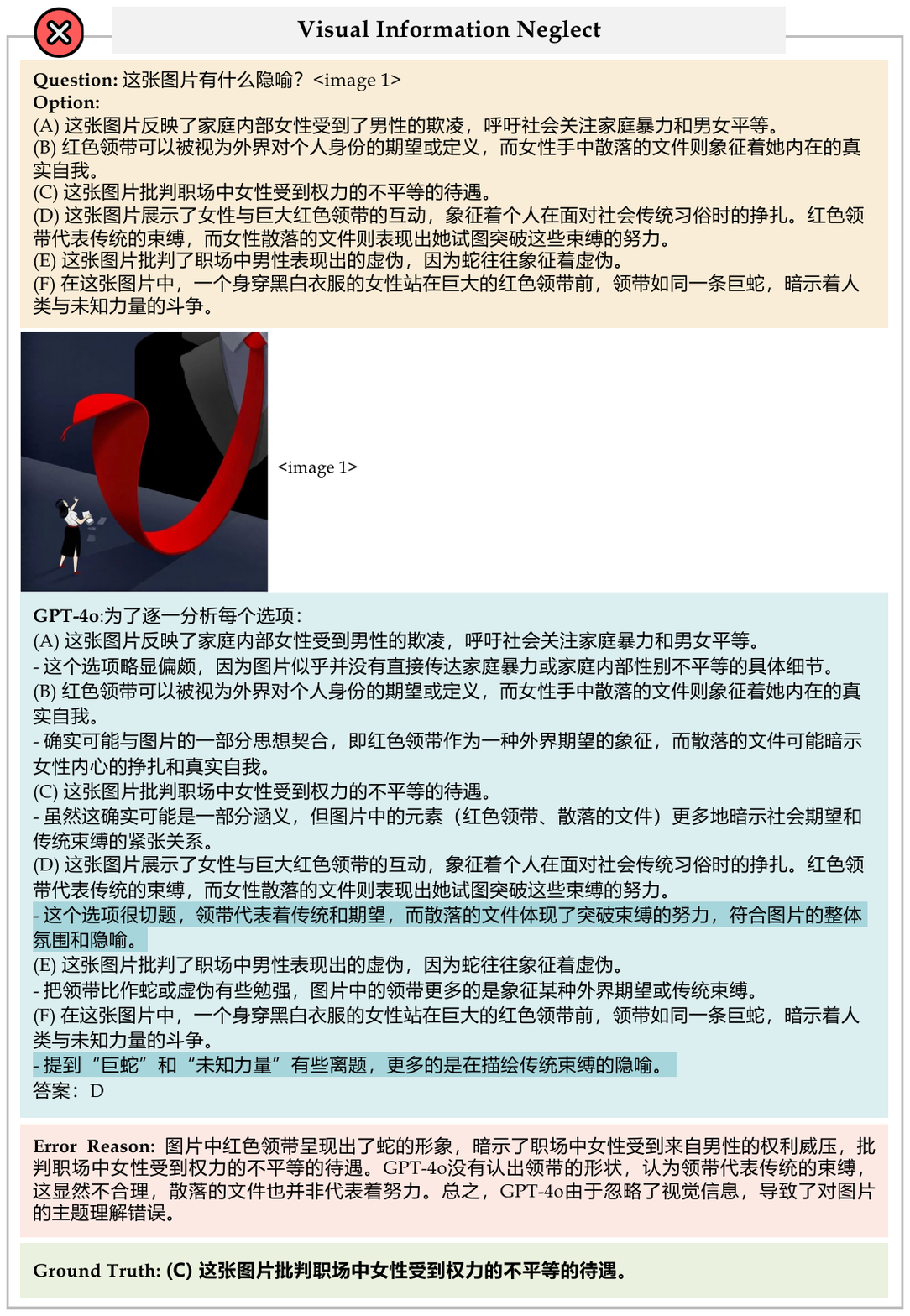}{Error 2: Visual Information Neglect}{A sample error case of \textit{Information Neglect}.}{fig:case_study_2}

\casestudyfigure{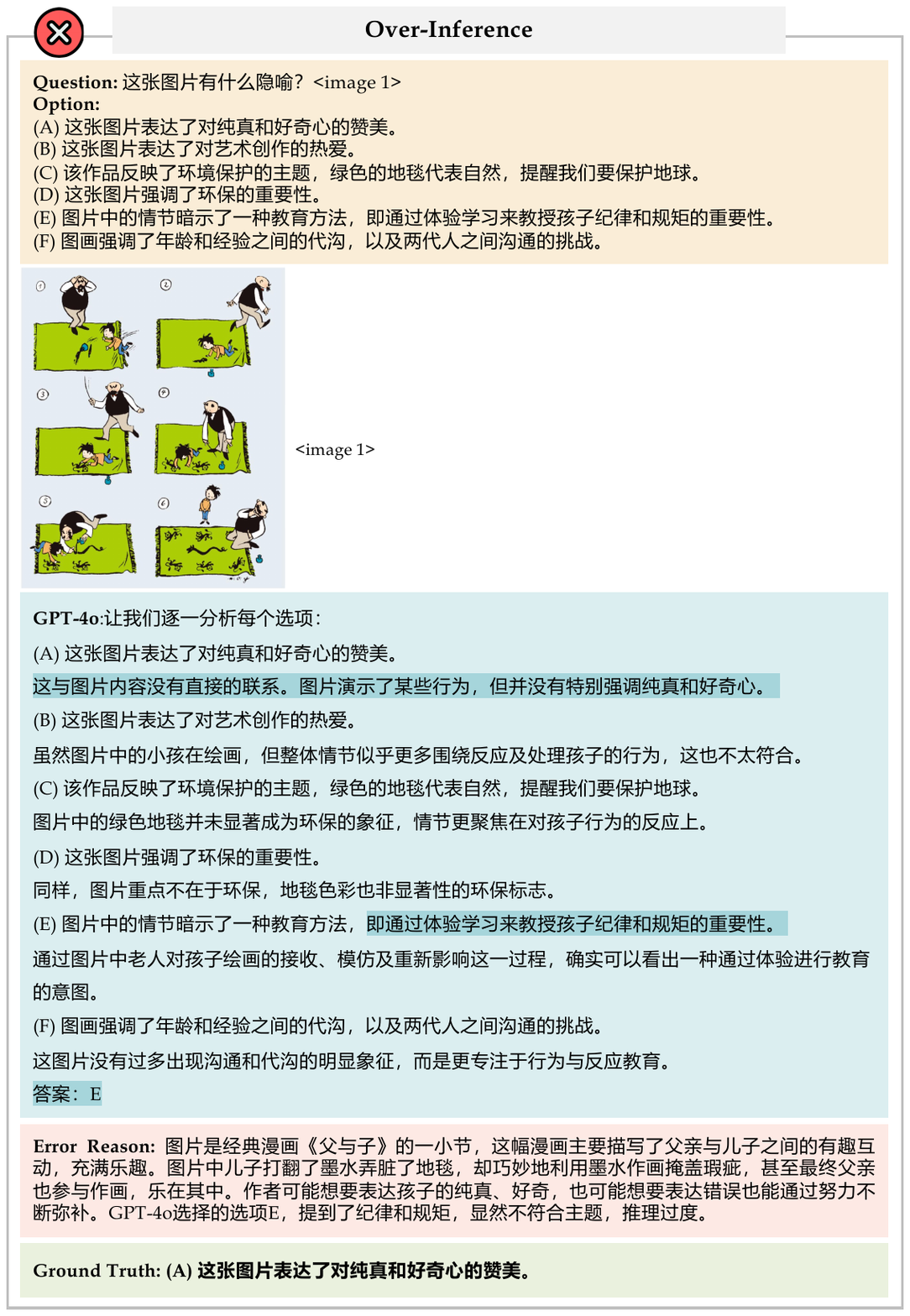}{Error 3: Over-Inference}{A sample error case of \textit{Over-Inference}.}{fig:case_study_3}

\casestudyfigure{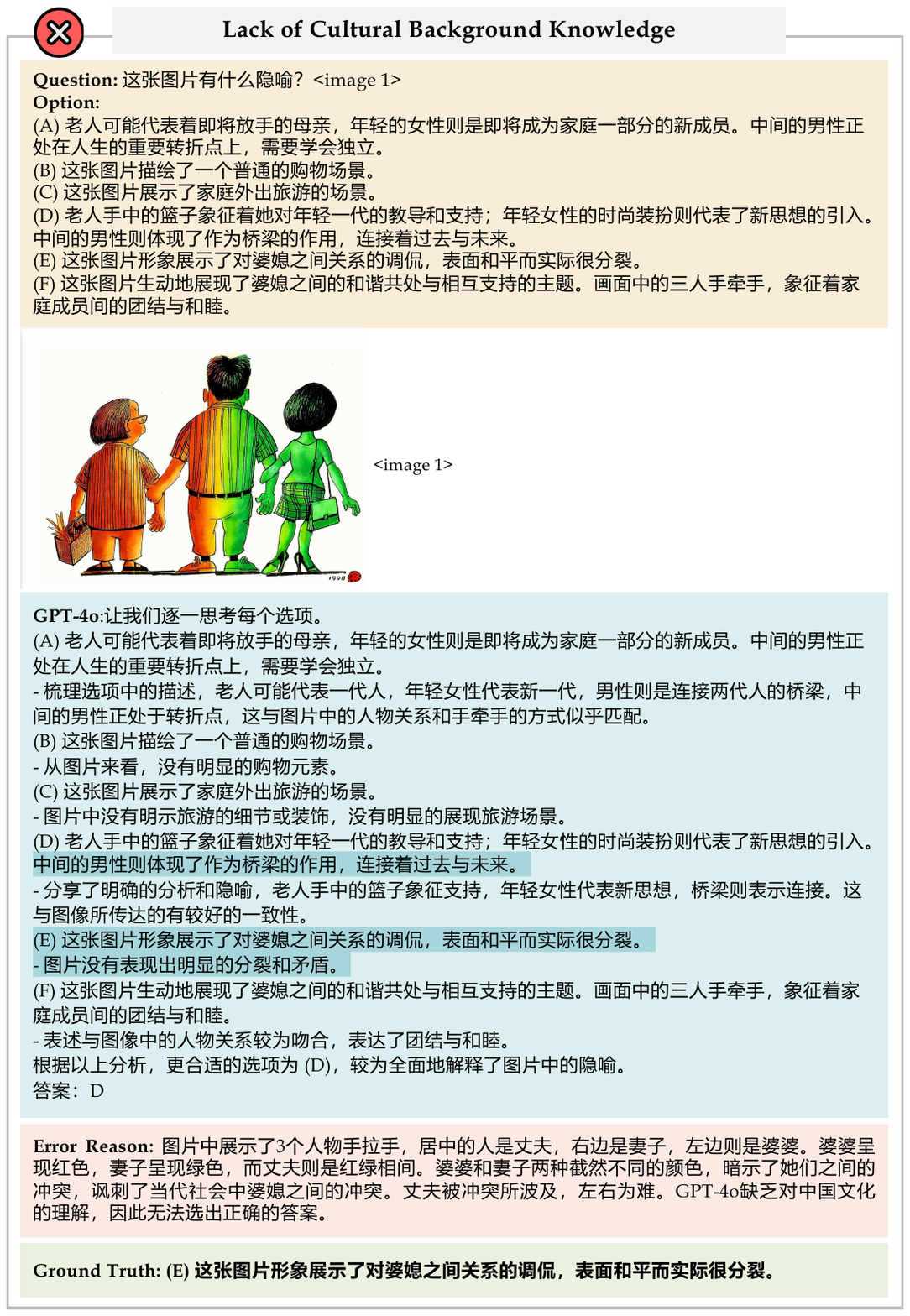}{Error 4: Lack of Cultural Background Knowledge}{A sample error case of \textit{Lack of Cultural Background Knowledge}.}{fig:case_study_4}

\casestudyfigure{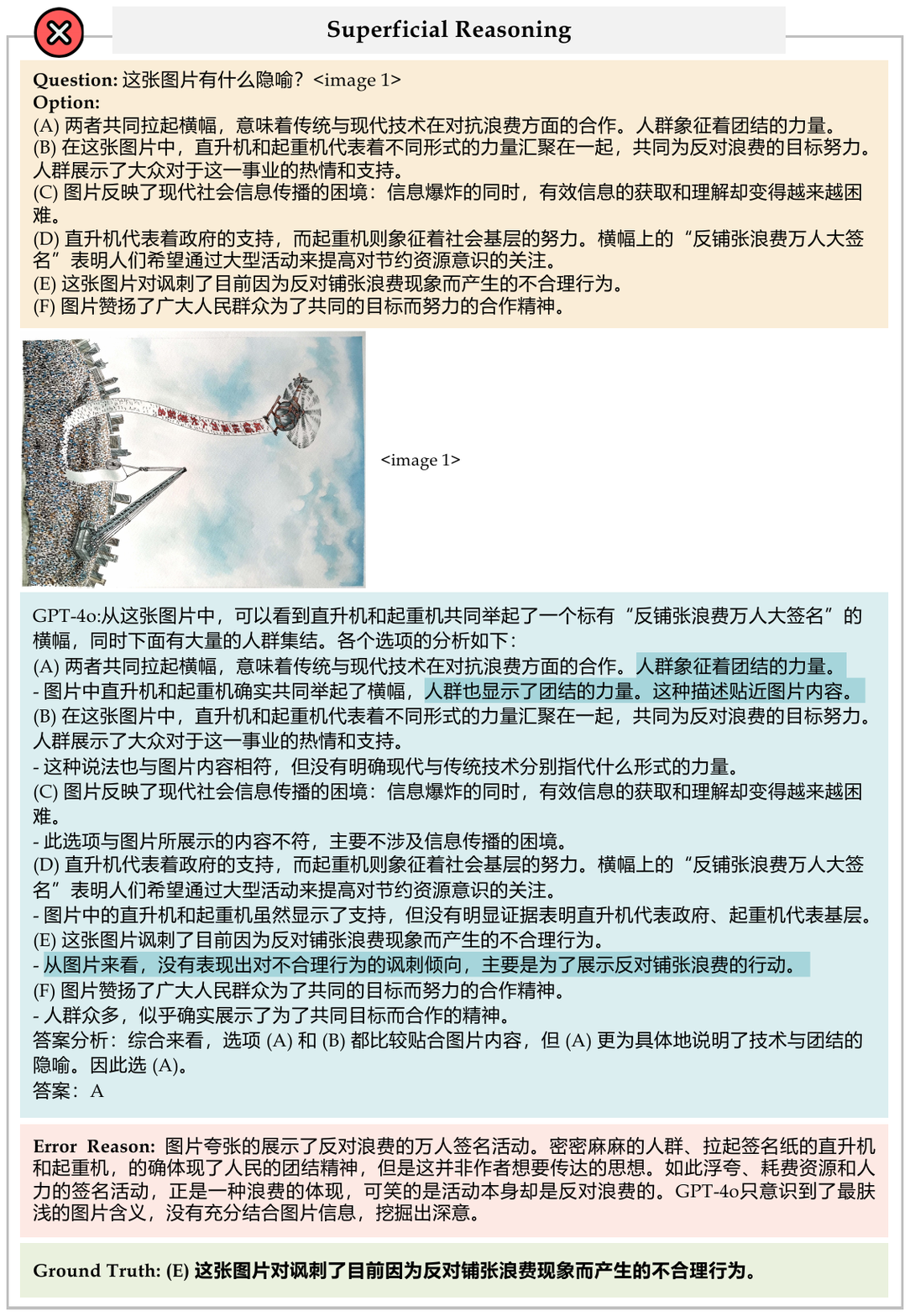}{Error 5: Superficial Reasoning}{A sample error case of \textit{Superficial Reasoning}.}{fig:case_study_5}

\casestudyfigure{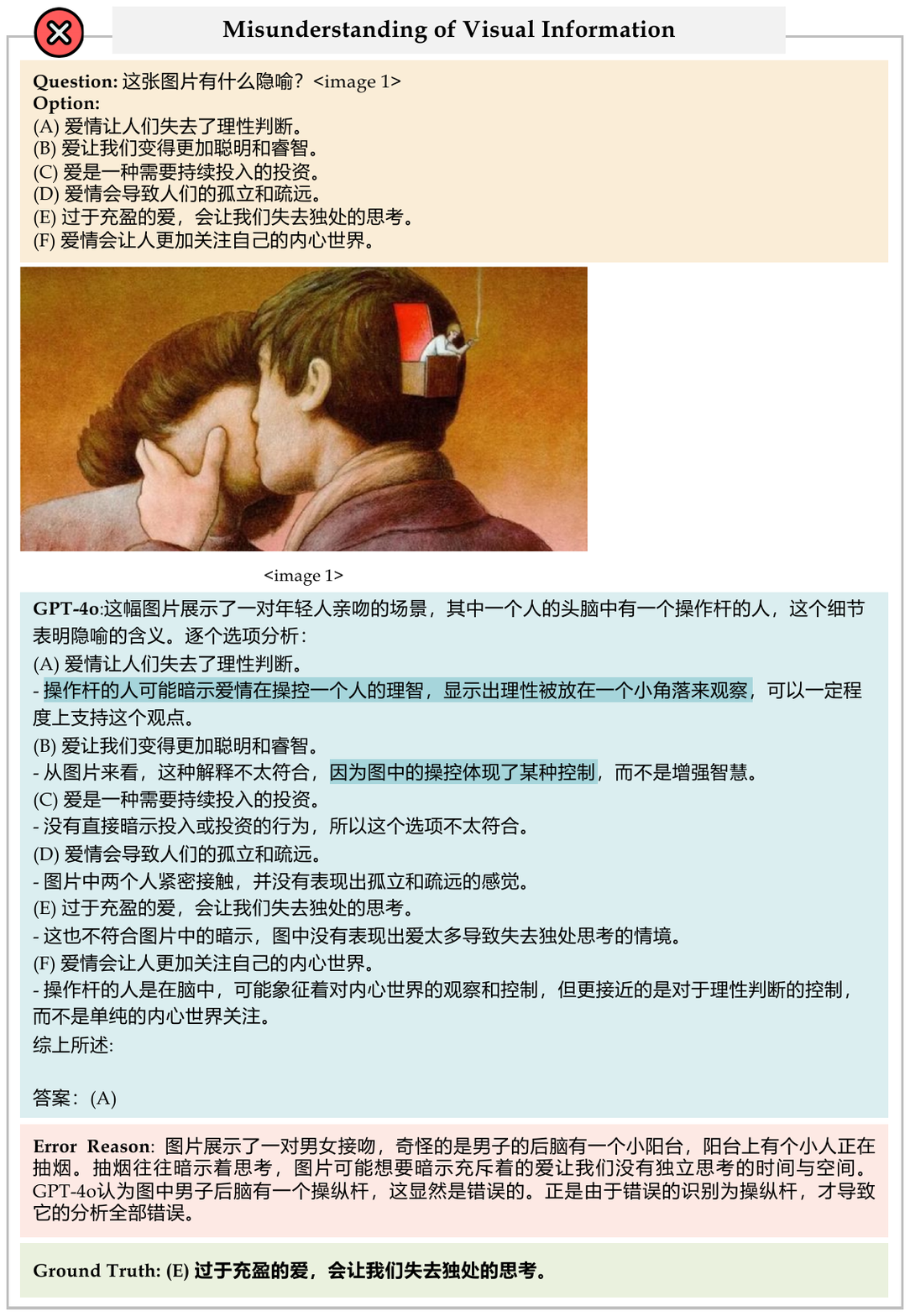}{Error 6: Misunderstanding of Visual Information}{A sample error case of \textit{Misunderstanding of Visual Information}.}{fig:case_study_6}